\documentclass[10pt,twocolumn,letterpaper]{article}

\usepackage[pagenumbers]{cvpr}
\definecolor{cvprblue}{rgb}{0.21,0.49,0.74}
\usepackage[pagebackref,breaklinks,colorlinks,allcolors=cvprblue]{hyperref}
\usepackage{microtype}
\usepackage{graphicx}
\usepackage{booktabs} 
\usepackage{multirow}
\usepackage{colortbl} 
\usepackage{xcolor}
\usepackage{array}
\usepackage{dsfont}
\usepackage{wrapfig}
\usepackage{enumitem}
\usepackage{mathrsfs}
\usepackage{amssymb}
\usepackage{amsmath}

\usepackage[utf8]{inputenc} 
\usepackage[T1]{fontenc}    
\usepackage{hyperref}       
\usepackage{url}            
\usepackage{booktabs}       
\usepackage{amsfonts}       
\usepackage{nicefrac}       
\usepackage{microtype}      
\usepackage{xcolor}         
\usepackage{booktabs}
\usepackage{multirow}
\usepackage{setspace}
\usepackage{graphicx} 
\usepackage{array}
\usepackage{tabularx}
\usepackage{makecell}
\usepackage{subcaption}

\usepackage{wrapfig}

\usepackage{tikz}
\usetikzlibrary{shapes.arrows}
\DeclareRobustCommand{\hollowarrow}{%
  \mathrel{%
    \tikz[baseline=-0.25em]{%
      \node[single arrow, draw=black, fill=white, inner sep=0pt, 
            minimum width=0.7em, minimum height=1.0em, 
            single arrow head extend=0.06em, line width=0.1em] {};%
    }%
  }%
}

\def\paperID{8507} 
\def\confName{CVPR}
\def\confYear{2026}

\title{Chain-of-Models Pre-Training: Rethinking Training Acceleration of \\ Vision Foundation Models}

\author{Jiawei Fan$^1$, Shigeng Wang$^1$, Chao Li$^1$, Xiaolong Liu$^2$, Anbang Yao$^{1}$\textsuperscript{\dag}\\
$^1$Intel Labs China, \ $^2$iMotion Automotive Technology\\
{\tt\small \{jiawei.fan, shigeng.wang, chao3.li, anbang.yao\}@intel.com}
}

\begin{document}
\maketitle
\renewcommand{\thefootnote}{\fnsymbol{footnote}} 
\footnotetext[2]{\ Corresponding author} 
\renewcommand{\thefootnote}{\arabic{footnote}} 

\begin{abstract}
In this paper, we present Chain-of-Models Pre-Training (CoM-PT), a novel performance-lossless training acceleration method for vision foundation models (VFMs). 
This approach fundamentally differs from existing acceleration methods in its core motivation: rather than optimizing each model individually, CoM-PT is designed to accelerate the training pipeline at the model family level, scaling efficiently as the model family expands.  
Specifically, CoM-PT establishes a pre-training sequence for the model family, arranged in ascending order of model size, called model chain.
In this chain, only the smallest model undergoes standard individual pre-training, while the other models are efficiently trained through sequential inverse knowledge transfer from their smaller predecessors by jointly reusing the knowledge in the parameter space and the feature space. As a result, CoM-PT enables all models to achieve performance that is mostly superior to standard individual training while significantly reducing training cost, and this is extensively validated across \textbf{45} datasets spanning zero-shot and fine-tuning tasks.
Notably, its efficient scaling property yields a remarkable phenomenon: training more models even results in higher efficiency. For instance, when pre-training on CC3M: i) given ViT-L as the largest model, progressively prepending smaller models to the model chain reduces computational complexity by up to 72\%; ii) within a fixed model size range, as the VFM family scales across 3, 4, and 7 models, the acceleration ratio of CoM-PT exhibits a striking leap: from 4.13$\times$ to 5.68$\times$ and 7.09$\times$.
Since CoM-PT is naturally agnostic to specific pre-training paradigms, we open-source the code\footnote{\ Code is available at \textit{https://github.com/deep-optimization/CoM-PT}} to spur further extensions in more computationally intensive scenarios, such as large language model pre-training.

\end{abstract}

\vspace{-1em}
\section{Introduction}
\label{sec:intro}

The success of vision foundation models (VFMs) is largely fueled by the advancement of the pre-training paradigm. As a representative example, the emergence of contrastive language-image pre-training (CLIP) \cite{radford2021learning} enables vision transformers (ViTs) to achieve state-of-the-art performance in zero-shot tasks and solidifies them as the \textit{de facto} vision encoders \cite{zhai2023sigmoid} for 
multimodal large language models (MLLMs) \cite{zhu2023minigpt,bai2023qwen,liu2024improved}. 
However, this success comes at enormous computational costs: pre-training a single ViT-L/14 on LAION-2B \cite{schuhmann2022laion} requires $1.2 \times 10^{5}$ A100 GPU hours, translating to a financial cost exceeding \$500,000.

\begin{figure}
    \vskip -0.2 in
    \centering
    \includegraphics[width=0.49\textwidth]{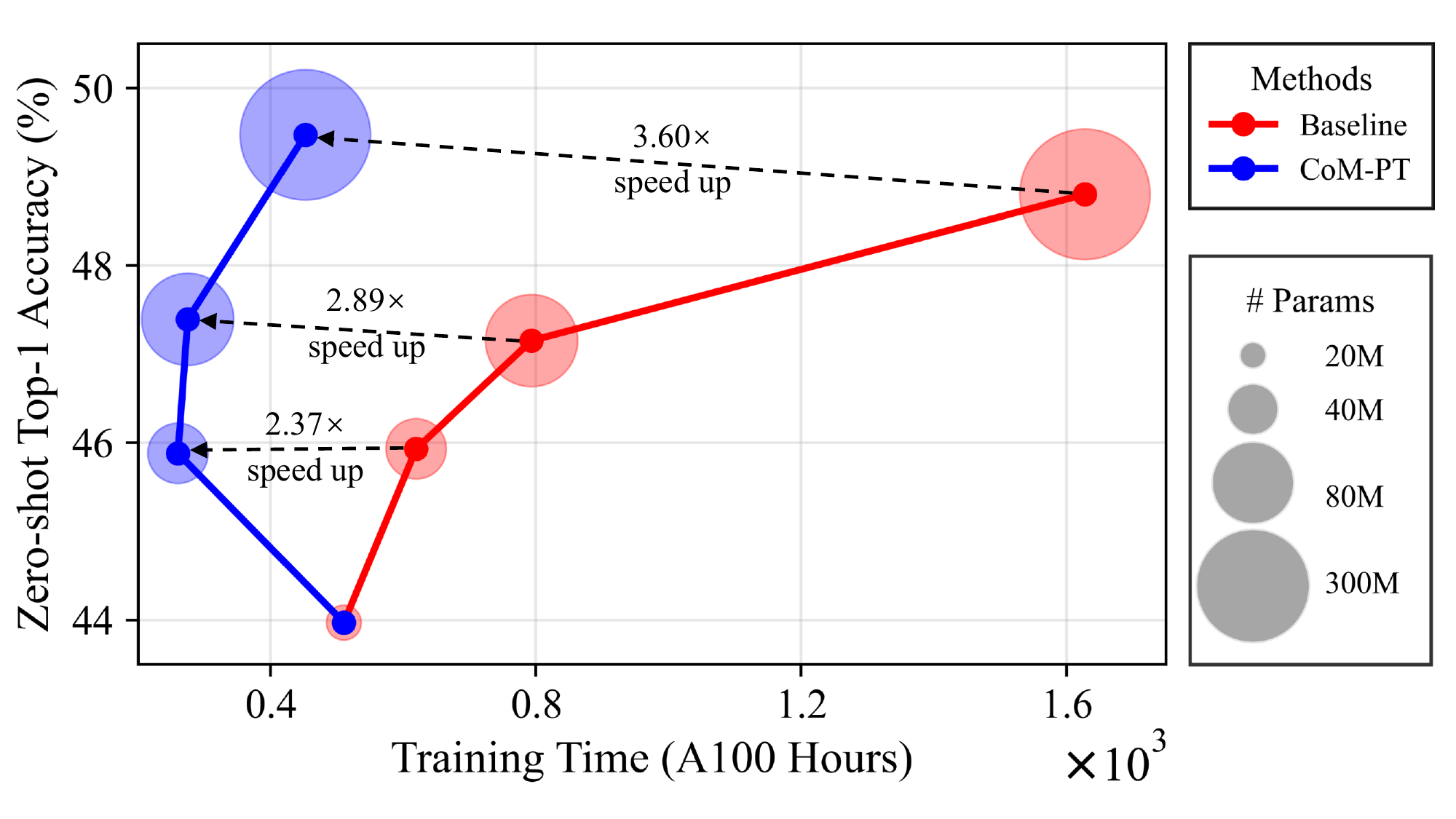}
    \vskip -0.1 in
    \caption{\textbf{Standard individual pre-training \textit{vs}. Chain-of-Models Pre-Training}. All ViTs \cite{dosovitskiy2020image} are trained on a combined dataset of CC3M \cite{sharma2018conceptual} and CC12M \cite{changpinyo2021cc12m}, and evaluated on ImageNet-1K \cite{deng2009imagenet}.}
    \label{fig:teaser}
    \vskip -0.2 in
\end{figure}

To mitigate the huge cost, various training acceleration techniques have been extensively investigated. At the foundational level, mixed-precision training \cite{micikevicius2018mixed} and system-level optimizations \cite{rasley2020deepspeed, kesselheim2021juwels} for large-scale clusters become standard practices.
Building upon these, some recent works attempt to improve pre-training efficiency through mask image modeling \cite{li2023scaling}, input token number reduction \cite{huang2024accelerating}, loss optimization \cite{wei2024fastclip}, and sophisticated data-efficient designs \cite{lisupervision, fan2023improving, zheng2024dreamlip}. However, despite belonging to different research lines, all these works share a common underlying viewpoint—\textit{optimizing the training of individual models}.
\begin{figure*}[ht]
    \vskip -0.2 in
    \centering
    \includegraphics[width=0.9\textwidth]{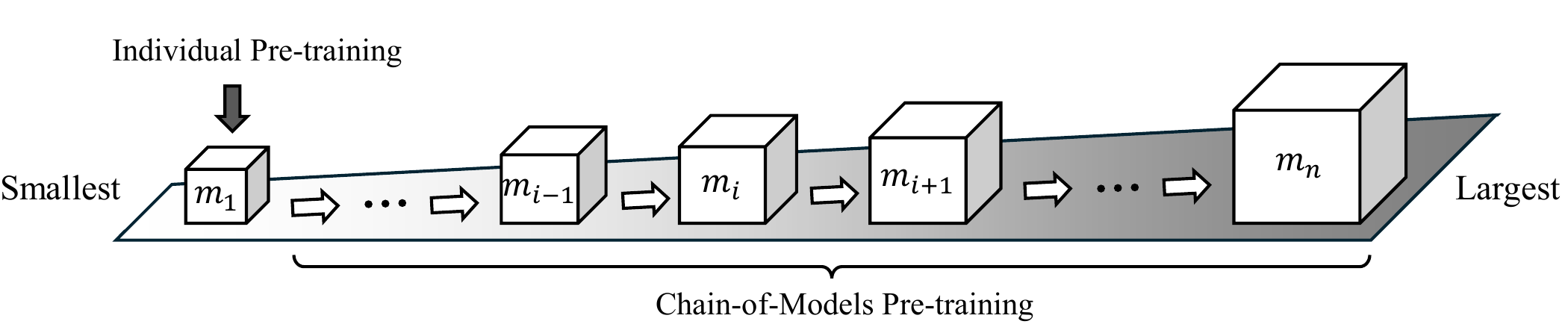}
    \vskip -0.1 in
    \caption{\textbf{Overview of Chain-of-Models Pre-training pipeline with inverse knowledge transfer relay.} Models are organized sequentially by ascending model size, from the smallest model $m_1$ to the largest $m_n$, forming a model chain. $m_1$ first undergoes standard individual pre-training. Each subsequent $m_i$ is then efficiently trained via inverse knowledge transfer (denoted as \ $\hollowarrow $) from its immediate predecessor $m_{i-1}$, forming a continuous relay along the model chain. Details of our proposed inverse knowledge transfer are in Figure \ref{fig:ikt}.}
    \label{fig:teaser1}
    \vskip -0.2 in
\end{figure*}

In contrast, we rethink the pre-training acceleration from a fundamentally different perspective. We notice that VFMs are typically pre-trained in the form of \textit{model family}, a series of architecturally similar models with varying sizes to accommodate different deployment scenarios.
As the field advances, the number of models in a VFM family (\textit{aka family size}) shows an ever-growing trend. This is mainly driven by two factors: i) the demand for specialized model sizes in emerging applications; ii) the broadening parameter range, driven by the development of increasingly larger models.
Inevitably, this trend creates a dilemma for pre-training model families: affording escalating pre-training costs from ever-growing family size or compromising deployment flexibility.
Faced with this dilemma, we question: \textbf{\textit{how can we achieve pre-training acceleration that scales efficiently with family size?}} To this end, we reexamine the standard pre-training pipeline and identify two primary bottlenecks:

\begin{itemize}[itemsep=1.5pt,topsep=0pt,parsep=0pt,leftmargin=14pt]
    \item[i)] \textbf{From the micro view, the main pre-training costs are sourced from large models in the model family.} In the classical ViT model family, the computational complexity expands around 4$\times$ when selecting a larger one. As reported in  \cite{cherti2023reproducible}, training the largest ViT-H/14 model costs over $80\times$ more than training ViT-B/32.
    \item[ii)] \textbf{From the macro view, the inefficiency stems from the individual training.} 
    Models are trained using the same optimization objectives, datasets, and training protocols, thus yielding common knowledge within the model family. However, with this fact ignored, traditional individual training is essentially highly repetitive and redundant.
\end{itemize}

We hypothesize that these two bottlenecks can be simultaneously addressed by redefining the training pipeline to enable knowledge reuse from small to large models within the model family.
Accordingly, we organize the models within a given family in ascending order of size to form a \textit{model chain}, a pre-training sequence that determines the overall pipeline structure. Starting with the smallest model pre-trained in the standard manner individually, each model in the chain leverages \textit{inverse knowledge transfer} to accelerate the pre-training of its successor, while maintaining performance comparable to individual training. 
This process proceeds sequentially along the model chain until all models are efficiently pre-trained. 
We refer to this training pipeline as \textbf{Chain-of-Models Pre-Training (CoM-PT)}.

In essence, the model chain serves as the core design of CoM-PT. 
By design, it eliminates the isolation of individual training by ensuring that each model inherits knowledge from its strongest predecessor, which in turn drastically reduces the prohibitive pre-training costs of larger models.
Beyond the efficiency gains along the chain, this design remarkably unlocks the training efficiency that scales favorably with the model family size. 
For instance, a counterintuitive phenomenon emerges in our experiments: the model chain structured as ViT-T$\rightarrow$ViT-S$\rightarrow$ViT-B$\rightarrow$ViT-L even reduces training costs by an additional 20\% on the CC3M dataset compared to the ViT-B$\rightarrow$ViT-L structure, while providing two additional models to support versatile deployment environments. This is because in the extended chain, ViT-B and ViT-S efficiently achieve convergence in significantly fewer epochs with assistance from their predecessors, leading to the total overhead of the first three models being even smaller than directly pre-training ViT-B from scratch. In principle, model chains can be either naturally formed based on existing model families or designed to align with specific requirements in practice. When opting for the latter case, maximizing training efficiency requires careful consideration of: i) selecting the optimal smallest model, and ii) determining scaling factors of model size between consecutive models, which are discussed in Section \ref{sec:principle}.

Conceptually, inverse knowledge transfer is designed to enable a seamless knowledge transfer relay along the model chain. 
In this process, each intermediate model serves first as a student learning from its predecessor, and then as a teacher guiding its larger successor. This differs from the traditional teacher-student learning paradigm in two key aspects: i) the relative size between teacher and student is inverted; 
ii) intermediate models serve dual roles to sustain the relay.
Under this context, we find that the effective relay hinges on comprehensively addressing different forms of knowledge representation.
We identify that the knowledge of a model is statically encoded in the parameter space, while exhibited dynamically in the feature space across different samples.
Therefore, we tailor weight initialization and feature distillation for knowledge reuse in the two spaces, respectively. The synergy between these two components ensures efficient knowledge transfer relay, yielding speedups that increase progressively along the chain, as shown in Figure \ref{fig:teaser}.

We empirically validate the effectiveness of our approach on CLIP. Extensive evaluation across 45 downstream datasets demonstrates that CoM-PT achieves significant acceleration while maintaining lossless performance.

\section{Related Works}

\subsection{Training Acceleration for VFMs}

VFMs have revolutionized the pre-training paradigm, demonstrating extraordinary scalability \cite{cherti2023reproducible, sun2023eva, sun2024eva} up to 18B parameters and 2B pre-training data scale.
The pre-training of large-scale VFMs relies on system-level optimization \cite{rasley2020deepspeed, kesselheim2021juwels} supported by many low-level techniques, such as distributed data parallel modules \cite{li2020pytorch}, mixed-precision training \cite{micikevicius2018mixed}, and memory
management \cite{rajbhandari2020zero}. 
Despite these advancements, the pre-training cost is still high, which motivates exploring its intrinsic mechanisms for achieving further acceleration.
Some works explore input token number reduction \cite{huang2024accelerating}, training of progressively activated weights \cite{pan2022budgeted} and frequency-space curriculum learning \cite{wang2023efficienttrain} to speed up the training process of single models.

In the context of accelerating CLIP \cite{radford2021learning}, FLIP \cite{li2023scaling} conducts random masking on image patches in CLIP, reducing computational complexity for each image and making it possible to train with a larger batch size. Another prominent research line is data-efficient CLIP, initially sourcing from studies that optimize CLIP learning objectives \cite{mu2022slip,yaofilip,gao2022pyramidclip}. DeCLIP \cite{lisupervision} first exploits data-efficient approaches by integrating three types of supervision, reducing the data scale from 400M to 88M without performance degradation. ALIP \cite{yang2023alip} pushes this boundary to 15M by effectively leveraging additional generated synthetic captions. Contemporaneously, LaCLIP \cite{fan2023improving} generates captions from LLM and treats them as text augmentation, while MLLM-A \cite{liu2023mllms} systematically studies the text augmentation from multiple MLLMs. DreamLIP \cite{zheng2024dreamlip} re-captures the image samples using stronger MLLMs and designs fine-grained contrastive loss, achieving significant performance improvements on the CC3M dataset. Follow-up works improve the performance by using long text captions \cite{wu2024lotlip} and self-distillation \cite{kim2025cosmos}.

\vspace{+0.4em}

\noindent \textit{In contrast, CoM-PT focuses on the accelerating the training pipeline at the model family level, with the novel objective of scaling efficiently with family size. }

\subsection{Knowledge Reuse between Models}

\label{sec:2.2}

\noindent \textbf{Weight Initialization from Models.} Many previous works have explored how to use the weights of well-trained models. Most studies explore how to transform \cite{ chen2024lemon, xia2024initializing, lin2020weight} or select  \cite{sanh2019distilbert, shleifer2020pre, trockman2023mimetic, xuinitializing} weights of large models to initialize small ones. In another research line, Net2Net \cite{chen2015net2net} first proposes reusing small model weights to initialize larger models via function-preserving transformations. NetExpand \cite{ding2023network} advances this approach by using SVD operations for width expansion and EMA updates for depth scaling. Similar works in LLM pre-training \cite{gong2019efficient, chen2022bert2bert, kim2023solar} explore this problem but are generally less effective, as verified in \cite{ding2023network}.

\vspace{+0.25em}

\noindent \textbf{Knowledge Distillation.} Knowledge distillation aims to utilize the learned knowledge of a well-trained teacher network to guide the training of a smaller student network. 
Pioneering vanilla KD \cite{hinton2015distilling} and FitNet \cite{romero2014fitnets} employ teacher logits and intermediate features for supervision, respectively.
Many works then explore different techniques to conduct distillation, such as feature transform \cite{yim2017gift, chen2021cross}, feature selection \cite{heo2019comprehensive, chen2021distilling}, feature ensemble \cite{chen2022improved,liunorm}, mask image modeling \cite{yang2022masked, huangmasked}. 
Another research line is collaborative learning between two networks.
Added with multi-branch classifiers, ONE \cite{zhu2018knowledge} formulates an online ensemble distillation scheme within a single network.
DML \cite{zhang2018deep} and follow-up works \cite{guo2020online, deng2022distpro, ypsilantis2024udon} explore the bidirectional online distillation between the two models. 

\vspace{+0.3em}

\noindent \textit{Inverse knowledge transfer tailors weight initialization and feature distillation for knowledge reuse in the parameter space and the feature space, respectively, serving the knowledge transfer relay along the model chain.}

\label{sec:related_works}

\section{Approach}

\label{sec:method}
In this section, we first formulate our chain-of-models pre-training and then apply it to CLIP as a \textbf{representative test case} to demonstrate how our method works in practice.

\subsection{Chain-of-Models Pre-training}

\label{sec:method_com}

\subsubsection{Basic Concepts}

\noindent \textbf{Model Chain.} For a model family $M$ consisting of $n$ models, let $m_i \in M$ represent the $i$-$th$ model in terms of model size, ordered from smallest to largest within the model family. The basic model chain of the model family $M$ is defined as $C_M :m_1 \rightarrow m_2 \rightarrow \cdots \rightarrow m_i \rightarrow \cdots \rightarrow m_n$, where right arrows indicate the direction of the pre-training sequence with the knowledge reuse of predecessors. As shown in Figure \ref{fig:teaser1}, the overall pipeline of CoM-PT mainly contains two steps. First, we pre-train $m_1$ individually using the standard protocol. Second, we utilize the learned knowledge of $m_1$ to assist the pre-training of $m_2$ via inverse knowledge transfer, sequentially conducting the same operation along the model chain, until the largest $m_n$ is pre-trained. This paradigm is more efficient than standard individual pre-training, because i) the overhead of pre-training $m_1$ is smallest across models in model family $M$, and ii) the knowledge reuse from $m_i$ accelerates the convergence of $m_{i+1}$ significantly.

\vspace{+0.25em}
\noindent\textbf{Inverse Knowledge Transfer.} Recall that to achieve effective knowledge transfer relay along the model chain, we outline weight initialization and feature distillation for knowledge reuse in the parameter space and the feature space, respectively. 
Since empirical results in Table \ref{tb:com_cmp_ind} reveal that the model chain itself drives the primary efficiency gains, we prioritize simplicity and implement these two components using vanilla approaches.
At any step from $m_i$ to larger $m_{i+1}$, these two components operate as follows.

\vspace{+0.25em}

\begin{figure}
    \centering
    \includegraphics[width=0.48\textwidth]{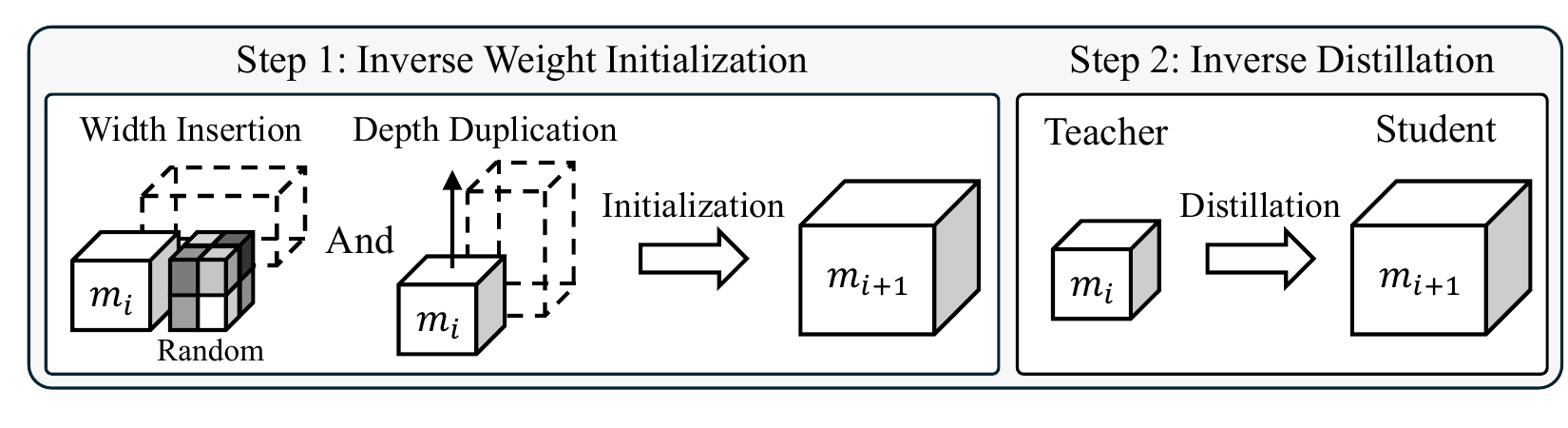}
    \vskip -0.15 in
    \caption{\textbf{Inverse knowledge transfer from $m_i$ to $m_{i+1}$.}}
    \label{fig:ikt}
    \vskip -0.20 in
\end{figure}

\noindent\textit{Inverse Weight Initialization.} 
The inverse weight initialization consists of two operations. As shown in Figure \ref{fig:ikt}, i) for block-block width differences, we directly insert the parameters of the small teacher into the large student, leaving the remaining parameters randomly initialized; ii) for layer-layer depth differences, we duplicate weights of each layer and index it as the succeeding layer. \textit{See Table \ref{exp:ablation_other} of Supplementary Materials for comparisons against sophisticated designs.}

\vspace{+0.25em}

\noindent\textit{Inverse Feature Distillation.} The inverse feature distillation has the same formulation as standard feature distillation. Let $F^t$$|$$F^s$ denote the output features of the teacher$|$student network. After applying the feature transform $\textbf{T}(\cdot)$ to the student feature to project it into the teacher's feature space, our inverse feature distillation is formulated as:
\begin{equation}
    \mathcal{L}_{IFD}(F^t, F^{s}) = \alpha||F^t - \textbf{T}(F^{s})||^2_2, 
    \label{eq1}
\end{equation}
where $\alpha$ is the balanced coefficient. Given any task loss $\mathcal{L}_{task}$ for pre-training, we typically ensure $\mathcal{L}_{IFD} < \mathcal{L}_{task}$ by setting an appropriate $\alpha$. \textit{We validate the rationale of this setting in Figure \ref{exp:abl_ifd} of Supplementary Materials}.

\subsubsection{Three Principled Rules for Model Chain Design}
\label{sec:principle}
Given a model chain $C_M$, we identify two operations that hold the potential to further improve the training efficiency. i) Adding a smaller $m_0$ before the first model $m_1$: compared to $m_1$, $m_0$ requires less training cost and can accelerate the convergence of $m_1$; ii) Inserting a medium-sized model $\hat{m}_{i}$ between any $m_i$ and $m_{i+1}$: the knowledge transfer relay by $\hat{m}_{i}$ can accelerate the convergence speed of $m_{i+1}$.
That said, the effectiveness of the two operations depends on whether the benefits outweigh the additional training costs from $m_0$ or $\hat{m}_{i}$. Furthermore, the allocation of training epochs along the model chain is tightly coupled with these structural modifications.
Therefore, we formulate three design principles to provide practical guidelines for model chain design.

\vspace{+0.25em}

\noindent\textbf{Optimal Smallest Model.} The smallest model size is selected based on the pre-training data scale. It should be sufficiently compact to maximize efficiency, while having proper capacity to fit the data distribution. Consequently, this guarantees lossless training acceleration (LTA) relative to standard individual pre-training and establishes a robust foundation for subsequent LTA steps along the chain.

\vspace{+0.25em}

\noindent\textbf{Choice of Intermediate Model Variants.}  These models are obtained by expanding the size of their respective teachers. We empirically find that using expansion ratios of VFM families ($2\times$ to $4\times$) yields considerable efficiency gains. The exact selection depends on the specific trade-off between computational cost and acceleration ratio. Specifically, larger factors (e.g., $4\times$) are opted to minimize the overall training cost, while smaller factors (e.g., $2\times$) are for maximizing the acceleration ratio.

\vspace{+0.25em}

\noindent\textbf{Training Epoch Allocation.} Training epochs allocation along the model chain decreases linearly as model size increases exponentially, under the premise of achieving LTA. Given an expected acceleration ratio, we first initialize the schedule by arranging the number of training epochs in a linearly decreasing manner along the model chain. The feasibility of this schedule is validated by the first network pair: if LTA is achieved, the target ratio is deemed plausible; otherwise, we relax the expected acceleration ratio and proportionally increase the training epochs of each model.

\vspace{+0.25em}

\noindent \textit{These three principled rules are verified in Section \ref{sec:exp_model_chain_design}}.

\subsection{Test Case: Applying CoM-PT to CLIP}

\label{sec:method_imp}
\noindent\textbf{Baseline Method and Task Loss.} To achieve convincing results, we select LaCLIP \cite{fan2023improving}, a strong data-efficient method of CLIP variants with a neat formulation, as the baseline method. We further boost its performance by using stronger text augmentation in \cite{zheng2024dreamlip}. Specifically, we expand each image-text pair to generate 16$\times$  more text variations and randomly select 4 subcaptions for each iteration update. Let $v_i$ and $t_i$ denote the image feature and text feature of the $i$-th pair among $N$ image-text pairs. With text augmentation, each image feature $v_i$ corresponds to $M$ augmented text features $t_{i,j}$, where $j=1, \dots, M$, forming $M$ positive pairs. The text-to-image loss is formulated as: 
\begin{equation}
    \mathcal{L}^{t2v} =  -
    \sum_{i=1}^N\ \sum_{j=1}^M \text{log}\frac{\text{exp}(\text{cos} \langle t_{i,j},v_i \rangle/\tau)}{\sum_{k=1}^N\text{exp}(\text{cos} \langle t_{i,j},v_k \rangle/\tau)}, 
\end{equation}
where $\tau$ and cos($\cdot$) are the temperature and cosine similarity, respectively. By duality of $\mathcal{L}^{t2v}$ and $\mathcal{L}^{v2t}$, the task loss is: 
\begin{equation}
    \mathcal{L}_{task} = (\mathcal{L}^{t2v} + \mathcal{L}^{v2t})/2.
\end{equation}

\vspace{+0.25em}

\noindent \textbf{Overall Loss.} Owing to its simplicity, CoM-PT can be applied to the baseline method without modification. Let $v^t, v^s$ and $t^t, t^s$ denote the visual and text features of the teacher and student, respectively. The distillation loss is defined as $\hat{\mathcal{L}}_{IFD} = (\mathcal{L}_{IFD}(v^t, v^{s}) + \mathcal{L}_{IFD}(t^t, t^{s}))/2$. Alongside inverse weight initialization, the overall training loss is:
\begin{equation}
    \mathcal{L} = \mathcal{L}_{task} + \hat{\mathcal{L}}_{IFD}.
    \label{eq4}
\end{equation}

\section{Experimental Setups}

\noindent \textbf{Pre-training Protocols.} We use CC3M \cite{sharma2018conceptual} and CC12M \cite{changpinyo2021cc12m} with text augmentation from \cite{zheng2024dreamlip}, combining them into Merged-15M. After augmentation, CC3M and Merged-15M contain 44.1M and 206.8M image-text pairs, respectively. Hyperparameter settings follow \cite{cherti2023reproducible, zheng2024dreamlip} and models within a family are trained with the same data.

\vspace{+0.25em}

\noindent \textbf{Evaluation Protocols.} We evaluate model performance across \textbf{45} datasets spanning zero-shot tasks and fine-tuning tasks. 
i) Zero-shot performance is assessed on ImageNet-1K \cite{deng2009imagenet} for the classification task, COCO \cite{lin2014microsoft} for the retrieval task, and VTAB+ \cite{schuhmann2022laion}, the largest zero-shot benchmark with 35 datasets, for testing domain transfer capabilities. ii) Fine-tuning performance is evaluated  on ADE-847/150 \cite{zhou2017scene}, PC-459/59 \cite{mottaghi2014role}, and VOC-20 \cite{everingham2011pascal} for open-vocabulary semantic segmentation tasks, and TextVQA \cite{singh2019towards}, ScienceQA \cite{lu2022learn}, POPE \cite{li2023evaluating}, and VQAv2 \cite{goyal2017making} for vision-language tasks.

\vspace{+0.25em}

\noindent \textbf{Model Family and Framework Selection.}  We verify our method's generalization ability on two model families: ViT \cite{dosovitskiy2020image} and Swin \cite{liu2021swin}. For downstream fine-tuning, we adopt SAN \cite{xu2023san} for open-vocabulary semantic segmentation tasks, and LLaVA-1.5-7B \cite{liu2024improved} for vision-language tasks.

\vspace{+0.25em}

\noindent \textbf{Baseline Method and Lossless Acceleration.} As we mentioned in Section \ref{sec:method_imp}, the baseline method is LaCLIP \cite{fan2023improving} with standard individual pre-training. 
We define lossless acceleration as achieving performance with less than 0.5\% accuracy loss compared to the baseline under the same training protocol while requiring reduced training time and computational complexity.
When all models in a model family satisfy this criterion, we consider lossless acceleration verified for the entire family. 
Training time is measured by NVIDIA A100 (80GB) GPUs, while the computational complexity consists of the forward, backward, and parameter update complexity for both models and optimizers.

\section{Experiments}

\label{sec:main experiments}

\begin{figure} 
    \includegraphics[width=0.45\textwidth]{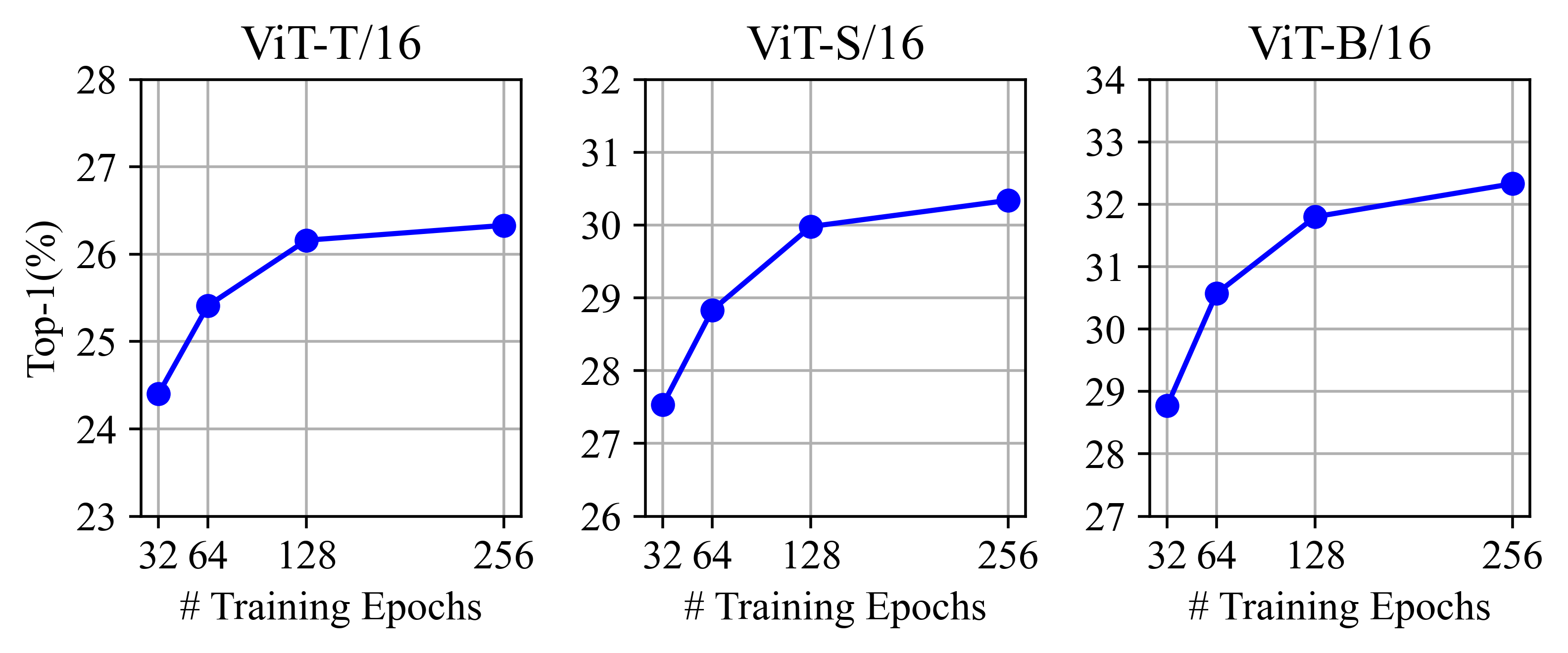}
    \vskip -0.15 in
    \caption{\textbf{Convergence analysis across different training epochs.} We train ViT-T/16, ViT-S/16, and ViT-B/16 on CC3M for 32, 64, 128, and 256 epochs to determine convergence points. Top-1 (\%) represents zero-shot classification accuracy on ImageNet-1K.}
    \vskip -0.2 in
    \label{fig:pilot_1}
\end{figure}

\subsection{Pilot Studies on Baseline Pre-training}
\label{sec:4.2}
Verifying performance-lossless acceleration requires two conditions for models trained with baseline: i) convergence, and ii) performance on par with prevalent counterparts.

\vspace{+0.25em}

\noindent \textbf{Standard Pre-training Epochs.} 
Early CLIP studies \cite{radford2021learning, cherti2023reproducible} standardize 32 epochs on large-scale datasets \cite{schuhmann2021laion, schuhmann2022laion}. Follow-up data-efficient CLIP methods  \cite{lisupervision,yang2023alip,fan2023improving,liu2023mllms,zheng2024dreamlip} directly adopt the same settings.
However, these works have largely overlooked that models trained on smaller datasets inherently require more epochs to converge. We empirically reveal that models remain far from convergence with only 32 epochs of training.
As illustrated in Figure \ref{fig:pilot_1}, our evaluation on CC3M across 32 to 256 epochs shows that performance steadily improves up to 128 epochs. Consequently, we establish 128 epochs as the standard baseline training epochs for CC3M.
For Merged-15M, we empirically set 64 epochs to account for the 5$\times$ increase in data scale.
\textit{Note that our method also achieves similar acceleration against baseline using 32 and 64 epochs, discussed in Section \ref{exp:ablation_setups}}.

\vspace{+0.25em}

\noindent \textbf{Comparison with Recent Data-efficient Method.} We select the original LaCLIP \cite{fan2023improving} and MLLM-A \cite{liu2023mllms} as counterparts, which represent the leading approaches that rely exclusively on contrastive loss formulations. To ensure a fair comparison, we exclude methods introducing auxiliary loss terms \cite{zheng2024dreamlip, wu2024lotlip}. As detailed in Table \ref{tab:polit_2}, our baseline outperforms both counterparts, establishing a solid foundation for validating the performance-lossless property of our method.

\begin{table}
\centering
\footnotesize
\setstretch{1.15}
\caption{\textbf{Performance comparison with recent top methods on the CC3M dataset.} Top-1 (\%) represents zero-shot classification accuracy on ImageNet-1K. \textit{Our method achieves similar acceleration with baseline using 32 epochs, demonstrated in Figure \ref{fig:different_epochs}}.}
\label{tab:polit_2}
\scriptsize
\begin{tabular}{l|c|c|
c}
\toprule
\textbf{Method} & \textbf{Text Augmentation Sources} & \textbf{\# Epochs} & \textbf{Top-1(\%)} \\
\midrule
CLIP & - & 32 & 16.00 \\ \midrule
LaCLIP & LLaMA \cite{touvron2023llama} & 32 & 21.50 \\ 
MLLM-A & Multi-MLLMs \cite{zhu2023minigpt,li2025otter,bai2023qwen,liu2024improved} & 32 & 25.00 \\ 
\midrule
Our Baseline & \multirow{2}{*}{ShareGPT4V \cite{chen2023sharegpt4v}} & 32 & 28.80 \\
(LaCLIP) &  & 128 & 31.80 \\
\bottomrule
\end{tabular}
\vskip -0.14 in
\end{table}

\begin{table*}[]
\centering
\setstretch{1.15}
\footnotesize

\caption{\textbf{Main results.} 
Evaluation is conducted on three benchmarks: ImageNet-1K, VTAB+, and COCO. Efficiency is measured in terms of training MACs and training time. Specifically, training MACs include all computations in the forward and backward propagations, as well as parameter updates, and training time denotes the required NVIDIA A100 (80GB) GPU hours for pre-training.
`Individual' and `accumulated' refer to specific model overhead and cumulative overhead along the model chain, respectively. 
Best results are \textbf{bolded}.
}
\vspace{-0.6em}
\resizebox{0.93\textwidth}{!}{%
\setlength{\tabcolsep}{4.6pt}
\begin{tabular}{@{}l|lc|ccc|cc|cc@{}}
\toprule
\rule{0pt}{1ex}
\multirow{3}{*}{\textbf{\ Model Family\ }} & \multirow{3}{*}{\textbf{Method}} & \multirow{3}{*}{\textbf{Model}} & \multicolumn{3}{c|}{\textbf{\ \ \ \ \ \ \ \ \   Zero-shot Top-1/R@1  (\%)\ \ \ \ \ \ \ \  \ }}  & \multicolumn{2}{c|}{\textbf{Training MACs ($10^{10}$G)}} & \multicolumn{2}{c}{\textbf{Training Time ($10^2$ GPU hrs)}} \\ \rule{0pt}{3ex}
                                  &                                    &                                 & ImageNet-1K& VTAB+ & COCO & Individual & Accumulated & Individual & Accumulated \\  \midrule  
\multicolumn{10}{c}{\textit{Pre-training Dataset: CC3M}} \\ \midrule
\multirow{8}{*}{\ \ ViT Family \ \ }        & \multirow{4}{*}{\textit{Baseline}} & ViT-T/16                        & 26.16 & 25.61 & 29.06 & 0.62 & 0.62 & 1.54 & 1.54 \\
                                  &                                    & ViT-S/16                        & 30.16 & \textbf{28.84} & 32.81 & 1.54 & 2.16 & 2.40 & 3.94 \\
                                  &                                    & ViT-B/16                        & 31.80 & \textbf{30.08} & 34.93 & 3.75 & 5.91 & 3.45 & 7.39 \\
                                  &                                    & ViT-L/16                        & 33.77 & 30.39 & 36.53 & 10.79 & 16.70 & 7.56 & 14.95 \\ \cmidrule(r){2-10}
                    & \multirow{4}{*}{CoM-PT}              & ViT-T/16                        & \textbf{26.16} & \textbf{25.61} & \textbf{29.06} & 0.62$_{(1.00\times)}$ & 0.62$_{(1.00\times)}$ & 1.54$_{(1.00\times)}$ & 1.54$_{(1.00\times)}$ \\
                                  &                                    & ViT-S/16                        & \textbf{30.24} & 28.61 & \textbf{34.15} & 0.32$_{(4.81\times)}$ & 0.94$_{(2.30\times)}$ & 0.52$_{(4.62\times)}$ & 2.06$_{(1.91\times)}$ \\
                                  &                                    & ViT-B/16                        & \textbf{31.83} & 29.89 & \textbf{35.67} & 0.59$_{(6.36\times)}$ & 1.53$_{(3.86\times)}$ & 0.54$_{(6.16\times)}$ & 2.62$_{(2.82\times)}$ \\ 
                                  &                                    & ViT-L/16                        & \textbf{34.27} & \textbf{30.78} & \textbf{39.18} & 1.41$_{(7.65\times)}$ & 2.94$_{(5.68\times)}$ & 1.02$_{(7.41\times)}$ & 3.64$_{(4.10\times)}$ \\ \midrule
\multirow{8}{*}{\ \ Swin Family \ \ }        & \multirow{4}{*}{\textit{Baseline}} & Swin-T                          & 33.84 & 28.72 & 37.29 & 0.94 & 0.94 & 2.22 & 2.22 \\
                                  &                                    & Swin-S                          & 35.88 & 30.55 & 39.46 & 1.97  & 2.91 & 3.29 & 5.51 \\
                                  &                                    & Swin-B                          & 36.13 & 31.17 & 39.84 & 3.48 & 6.39 & 3.73 & 9.24 \\
                                  &                                    & Swin-L                          & 36.77 & 31.80 & 40.23 & 7.86 & 14.25 & 6.76 & 16.00 \\ \cmidrule(r){2-10}
              & \multirow{4}{*}{CoM-PT}              & Swin-T                          & \textbf{33.84} & \textbf{28.72} & \textbf{37.29} & 0.94$_{(1.00\times)}$ & 0.94$_{(1.00\times)}$ & 2.22$_{(1.00\times)}$ & 2.22$_{(1.00\times)}$ \\
                                  &                                    & Swin-S                          & \textbf{36.19} & \textbf{31.35} & \textbf{40.06} & 0.43$_{(4.58\times)}$ & 1.37$_{(2.12\times)}$ & 0.68$_{(4.83\times)}$ & 2.90$_{(1.90\times)}$ \\
                                  &                                    & Swin-B                          & \textbf{36.71} & \textbf{32.53} & \textbf{41.06} & 0.52$_{(6.69\times)}$ & 1.89$_{(3.38\times)}$ & 0.58$_{(6.43\times)}$ & 3.48$_{(2.66\times)}$ \\ 
                                  &                                    & Swin-L                          & \textbf{37.06} & \textbf{32.68} & \textbf{41.95} & 0.84$_{(9.35\times)}$ & 2.73$_{(5.22\times)}$ & 0.73$_{(9.26\times)}$ & 4.21$_{(3.80\times)}$ \\ \midrule
\multicolumn{10}{c}{\textit{Pre-training Dataset: Merged-15M}} \\ \midrule
\multirow{8}{*}{\makecell{ViT Family}}        & \multirow{4}{*}{\textit{Baseline}} &  ViT-S/16                       & 43.97 &  35.35 & 43.23 & 3.61 & 3.61 & 5.10 & 5.10 \\
                                  &                                    & ViT-M/16                         & \textbf{45.89} & 34.97  & \textbf{44.87}  & 5.52 & 9.13 & 6.20 & 11.30 \\ 
                                  &                                    & ViT-B/16                         & 47.13 &  \textbf{36.98} & 46.44  & 8.79 & 17.92 & 7.94 & 19.24 \\
                                  &                                    & ViT-L/16                         & 48.88 &  \textbf{38.90} & 48.03  & 25.33 & 43.25 & 16.28 & 35.52 \\ \cmidrule(r){2-10}
          & \multirow{4}{*}{CoM-PT}              &  ViT-S/16                       & \textbf{43.97} &  \textbf{35.35} & \textbf{43.23} & 3.61$_{(1.00\times)}$ & 3.61$_{(1.00\times)}$  & 5.10$_{(1.00\times)}$ & 5.10$_{(1.00\times)}$ \\
                                  &                                    & ViT-M/16                         & 45.85 &  \textbf{35.22} & 44.83  & 2.19$_{(2.52\times)}$ & 5.80$_{(1.57\times)}$ & 2.60$_{(2.38\times)}$ & 7.70$_{(1.46\times)}$ \\ 
                                  &                                    & ViT-B/16                         & \textbf{47.15} &  36.89 & \textbf{46.50}  & 2.82$_{(3.12\times)}$ & 8.62$_{(2.08\times)}$ & 2.75$_{(2.89\times)}$ & 10.45$_{(1.84\times)}$ \\ 
                                  &                                    & ViT-L/16                         & \textbf{49.03} &  38.89 & \textbf{48.79}  & 6.62$_{(3.83\times)}$ & 15.24$_{(2.84\times)}$ & 4.52$_{(3.60\times)}$ & 14.97$_{(2.37\times)}$ \\ \bottomrule
\end{tabular}
}
\vskip -0.05 in
\label{tab:main}
\end{table*}

\subsection{Experiments on Standard VFM Families}
\label{sec:main_result}
We then pre-train ViT and Swin families on the CC3M dataset, and the ViT family on the Merged-15M dataset.

\subsubsection{Zero-shot Classification and Retrieval}
After pre-training, CLIP models are capable of performing zero-shot classification and retrieval tasks directly. In Table \ref{tab:main}, we present the zero-shot performance on ImageNet-1K, VTAB+, and COCO, along with the corresponding training overhead in terms of MACs and GPU hours.

\vspace{+0.25em}
\noindent \textbf{Performance on Three Benchmarks.} The results demonstrate that our method achieves lossless acceleration at both model and family levels. 
i) Models pre-trained with CoM-PT generally exceed the baseline on three benchmarks, especially for the largest model. 
Taking Swin-L on the CC3M dataset as an example, CoM-PT surpasses the baseline with a notable margin of 0.29\%$|$0.88\%$|$1.72\% on ImageNet-1K$|$VTAB+$|$COCO benchmark.
ii)  In the rare cases where performance slightly falls below baseline, differences remain within the 0.5\% lossless threshold, ensuring family-level lossless acceleration. \noindent \textit{Detailed results on VTAB+ benchmark are shown in Table \ref{tb:vtab+} of Supplementary Materials.}

\begin{figure}
    \centering
    \includegraphics[width=0.44\textwidth]{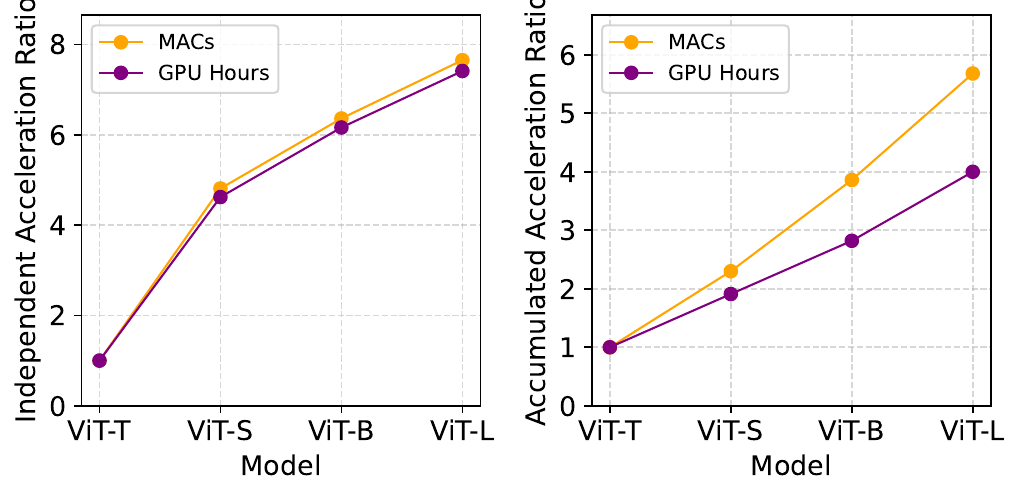}
    \vskip -0.12 in
    \caption{\textbf{Comparison of acceleration ratios, measured by two efficiency metrics.} We report both training MACs and A100 GPU hours for the ViT family pre-trained on the CC3M dataset.}
    \vskip -0.2 in
    \label{fig:ratios_compare}
\end{figure}

\vspace{+0.25em}
\noindent \textbf{Training Acceleration in Terms of MACs and GPU Hours.} Compared to the baseline method, our CoM-PT shows significant reductions in pre-training cost. First, the MACs and GPU hours are consistently high from the view of the individual acceleration ratio, as shown in Figure \ref{fig:ratios_compare} (left). \textit{With the increase of model size , the acceleration ratio becomes higher}, up to around $7.5\times$ at ViT-L. Benefiting from this attribute, the total cost for the entire model family is lower than that of training the largest model independently across three sets of experiments. Second, as shown in Figure \ref{fig:ratios_compare} (right), the accumulated acceleration ratio in GPU hours is slightly lower than that in MACs.
This occurs because, as model size decreases, GPU utilization decreases correspondingly, resulting in disproportionately longer training time for smaller models. Even with this, as illustrated in Table \ref{tab:main}, CoM-PT still achieves 4.10$\times$$|$3.80$\times$ acceleration for the ViT$|$Swin family on CC3M and 2.37$\times$ acceleration for the ViT family on Merged-15M.

\begin{table*}[]
    \vskip -0.02 in
    \centering
    \setstretch{1.15}
    \scriptsize
    \caption{\textbf{Fine-tuning on downstream tasks.} With backbones pre-trained by baseline method and CoM-PT, the framework SAN \cite{xu2023san} for open-vocabulary semantic segmentation is fine-tuned on the COCO-Stuff \cite{caesar2018coco} training set, while the LLaVA-1.5-7B \cite{liu2024improved} for vision-language tasks undergoes LoRA-based  visual instruction fine-tuning following its official protocol. Both ViT-B/16 and ViT-L/16 are pre-trained on the CC3M dataset. The `PT method' refers to the pre-training method of the backbone. Best results are \textbf{bolded}.}
    \vspace{-0.6em}
    \begin{tabular*}{0.93\textwidth}{@{\extracolsep{\fill}}l| l |c c c c c c |c c c c c@{}}
        \toprule
        & & \multicolumn{6}{c|}{\textbf{Open-Vocabulary Semantic Segmentation Tasks}} & \multicolumn{5}{c}{\textbf{Vision-Language Tasks}} \\ 
        \textbf{Backbone} & \textbf{PT Method} & \rotatebox{90}{ADE-847} & \rotatebox{90}{ADE-150} & \rotatebox{90}{PC-459} & \rotatebox{90}{PC-59} & \rotatebox{90}{VOC} & \rotatebox{90}{Avg.} & \rotatebox{90}{TextVQA} & \rotatebox{90}{ScienceQA  } & \rotatebox{90}{POPE} & \rotatebox{90}{VQAv2} & \rotatebox{90}{Avg.} \\
        \midrule
        \multirow{2}{*}{ViT-B/16} & \textit{Baseline} & 0.45 & 11.40 & 1.07 & 36.12 & 68.59 & 23.53 & 43.08 & 62.57 & 73.62 & 54.16 & 58.36 \\
              & CoM-PT      & \textbf{3.18} & \textbf{18.30} & \textbf{6.68} & \textbf{40.73} & \textbf{76.96} & \textbf{29.17} & \textbf{43.21} & \textbf{63.41} & \textbf{74.17} & \textbf{54.59} & \textbf{58.85} \\ \midrule
        \multirow{2}{*}{ViT-L/16} & \textit{Baseline} & 1.03 & 13.58 & 4.47 & 33.99 & 69.76 & 24.57 & 43.64 & \textbf{63.56} & 74.50 & 55.68 & 59.35 \\
              & CoM-PT      & \textbf{4.73} & \textbf{20.14} & \textbf{7.00} & \textbf{42.17} & \textbf{79.52} & \textbf{30.71} & \textbf{43.83} & 63.51 & \textbf{75.15} & \textbf{56.07} & \textbf{59.64} \\ 
        \bottomrule
    \end{tabular*}
    \vspace{-5mm}
    \label{tab:transfer}
\end{table*}

\subsubsection{Fine-tuning on Downstream Tasks}
As shown in Table \ref{tab:transfer}, we further evaluate the transferability of CoM-PT by fine-tuning on two types of downstream tasks.

\vspace{+0.25em}
\noindent \textbf{Open-Vocabulary Semantic Segmentation Tasks.}
Overall, models trained with CoM-PT outperform baselines across all five datasets, with average improvements of 5.64\%$|$6.14\% for ViT-B/16$|$ViT-L/16. For fine-grained datasets ADE-847 and PC-459, the gain from CoM exceeds 3$\times$ over the baseline, suggesting the acceleration stems from both faster convergence and better representation capability.

\vspace{+0.25em}
\noindent\textbf{Vision-Language Tasks.} Vision encoders pre-trained with CoM-PT achieve highly competitive performance against baselines across four datasets. It outperforms baseline in text understanding, question answering, and hallucination handling on TextVQA, VQAv2, and POPE, while showing only a negligible 0.05\% performance drop on ScienceQA.

\vspace{+0.25em}
\noindent \textit{These findings validate that CoM-PT accelerates VFM pre-training without compromising the transferability of models.}

\subsection{Empirical Studies on Model Chain Design}

\begin{figure}
    \centering
    \includegraphics[width=0.44\textwidth]{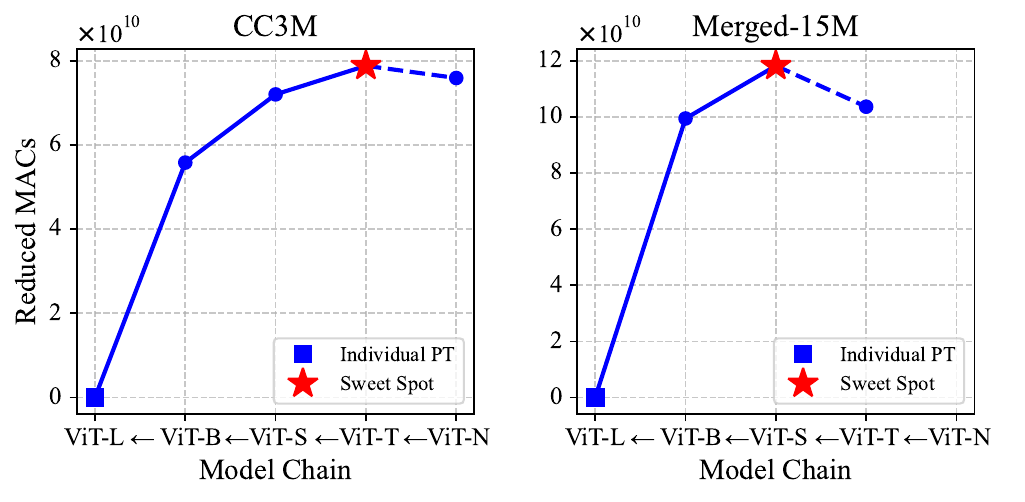}
    \vskip -0.10 in
    \caption{\textbf{Reduced training MACs \textit{vs.} model chain starting with smaller models.} The reduced MACs represent the total training cost of the
    model chain relative to the cost of individually pre-training ViT-L. All models are performance-lossless. }
    \vskip -0.22 in
    \label{fig:smallest model}
\end{figure}

\label{sec:exp_model_chain_design}
In Section \ref{sec:principle}, we propose three principled rules for model chain design to enable higher training efficiency of our CoM-PT. In this subsection, we verify these rules one by one.
\textit{Note that experiments are conducted on the ViT family with ViT-L/16 as the largest model, and all trained models are performance-lossless relative to the baseline pre-training.}

\subsubsection{Training Efficiency \textit{vs.} Model Family Size}
\label{sec:efficiency vs. size}
We begin by evaluating the first two rules, which dictate the structure of the model chain, including the specific model variants and family size. Crucially, verifying these rules serves to substantiate our core motivation: our method scales efficiently as the model family size expands.

\vspace{+0.25em}
\noindent\textbf{Longer Chain Starting with Smaller Models.}
We construct model chains by progressively adding smaller models, starting from solo ViT-L. For example, ViT-B$\rightarrow$ViT-L becomes ViT-S$\rightarrow$ViT-B$\rightarrow$ViT-L after one addition. 
Efficiency improvement is quantified by the MAC reduction relative to an individually pre-trained ViT-L. Results in Figure \ref{fig:smallest model} show that: i) the reduced MACs consistently increase until an optimal `sweet spot', then decline; ii) the optimal smallest model varies with data scale: ViT-T for CC3M and ViT-S for Merged-15M. 
Notably, the optimal configuration on CC3M yields up to a 72\% reduction in training MACs.
These results empirically verify the first rule and demonstrate that a longer model chain starting with a suitably small model can even reduce the overall training cost. 

\vspace{+0.25em}

\begin{figure}
    \centering
    \includegraphics[width=0.44\textwidth]{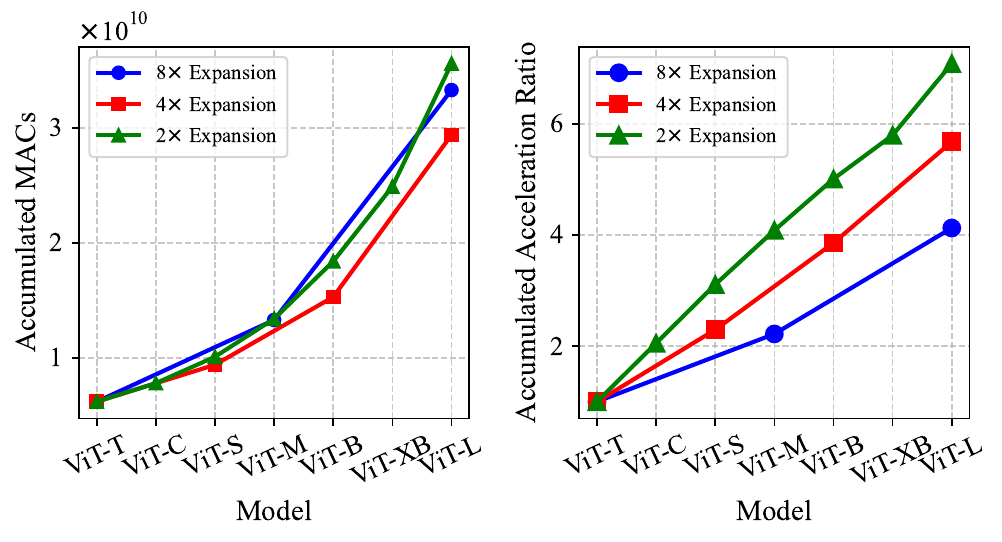}
    \vskip -0.10 in
    \caption{\textbf{Training MACs and acceleration ratios of model chains with different expansion ratios}. Experiments are conducted on the CC3M dataset. All models are performance-lossless.}
    \vskip -0.22 in
    \label{fig:best_interval}
\end{figure}

\vspace{+0.25em}
\noindent\textbf{Longer Chain with Smaller Model Size Intervals.}
With the identified optimal smallest model, we then study the optimal model size intervals. On the  CC3M dataset, we decrease the expansion ratio between two consecutive models from $8\times$ to $4\times$ and $2\times$, yielding three model chains comprising 3, 4, and 7 models, respectively. Results in Figure \ref{fig:best_interval} show that compared to the $8\times$ expansion:
i) the $4\times$ expansion reduces total training MACs by 7.3\%;
ii) the $2\times$ expansion boosts the accumulated acceleration ratio from 4.13 to 7.09;
iii) crucially, the $4\times$ and $2\times$ expansions provide 1 and 4 additional intermediate models, respectively.
These results verify the second rule and suggest that training a longer model chain with smaller model size intervals via CoM-PT maintains a near-constant overall training costs.

\vspace{+0.25em}

\noindent\textit{Together, the two evaluations validate our core motivation: CoM-PT scales highly efficiently with the increase of family size, yielding a larger VFM family for flexible deployment.}

\begin{figure} 
    \centering
    \vskip -0.15 in
    \includegraphics[width=0.40\textwidth]{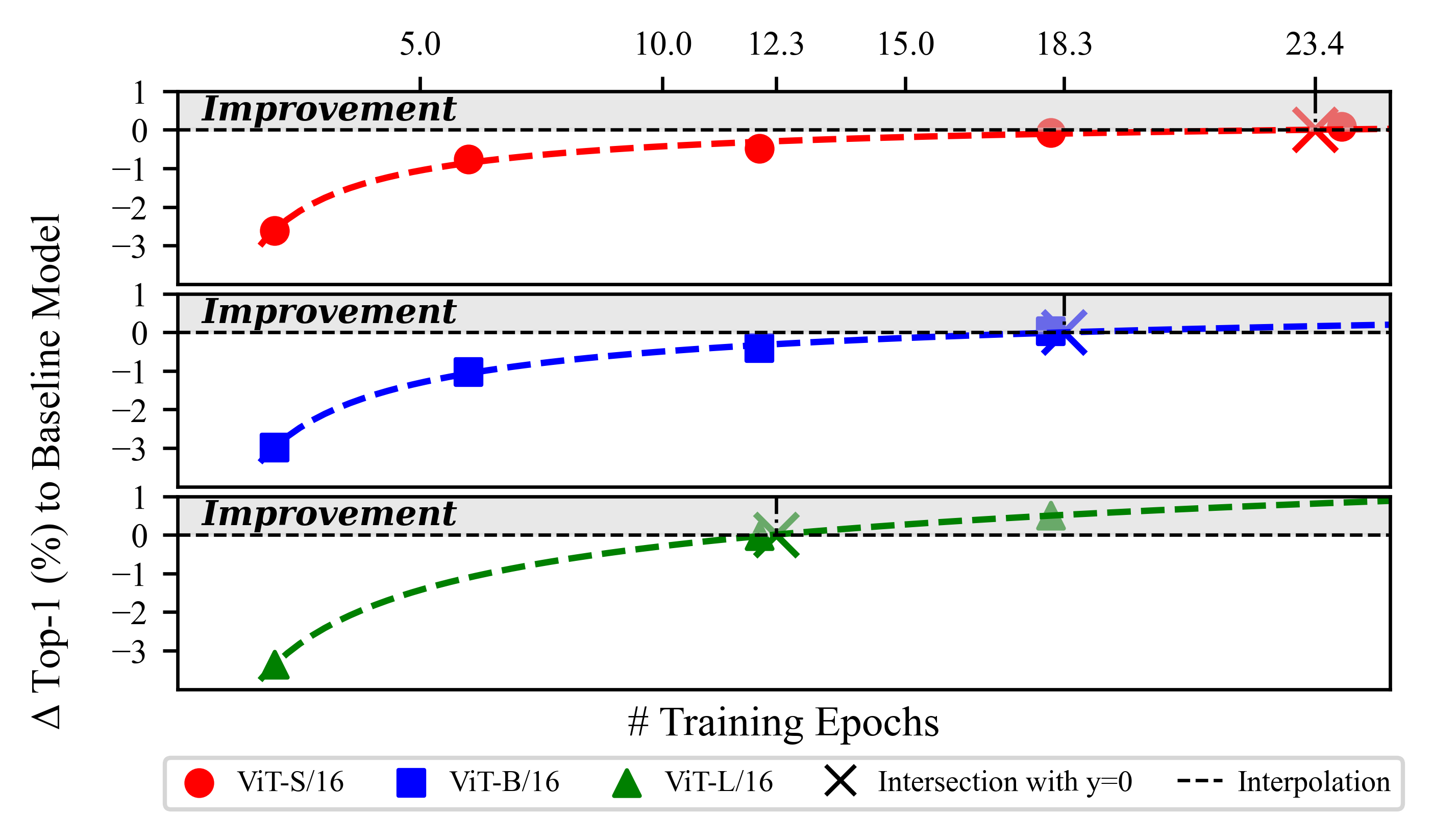}
    \vskip -0.1 in
    \caption{\textbf{Training epoch allocation.} Along model chain structured ViT-T$\rightarrow$ViT-S$\rightarrow$ViT-B$\rightarrow$ViT-L, we determine the minimum training epochs required per model to achieve lossless acceleration via CoM-PT, compared to standard 128-epoch baselines on CC3M.}
    \label{fig:exp_strategy}
    \vskip -0.22 in
\end{figure}

\begin{table*}[htbp]
    \centering
    \caption{\textbf{Ablation Studies.} (a) We study efficiency gain from the model chain, by comparing models progressively trained along the model chain (i.e., ViT-T$\rightarrow$ViT-S$\rightarrow$ViT-B) against those independently initialized from a trained ViT-T. (b) We study the complementarity of inverse weight initialization (IWI) and inverse feature distillation (IFD) using ViT-T$\rightarrow$ViT-S short chain. All models are trained on CC3M. }
    \vskip -0.1 in
    \scriptsize 
    \begin{subtable}{0.48\linewidth}
        \centering
        \setstretch{0.93}
        \caption{Efficiency gain from the model chain.}
        \label{tb:com_cmp_ind}
        \begin{tabular}{l|l|c|c|cc}
            \toprule
            \textbf{Model} & \textbf{Method} & \textbf{Init. From} & \textbf{\# Epochs} & \textbf{IN-1K (\%)} & \textbf{COCO (\%)} \\
            \midrule
            \multirow{3}{*}{ViT-S} 
            & Baseline & Random & 128 & 30.16 & 32.81 \\
            & IWI & ViT-T & 32  & 30.11 & 32.53 \\
            & CoM-PT & ViT-T & \textbf{24}  & \textbf{30.24} & \textbf{34.15} \\
            \midrule
            \multirow{3}{*}{ViT-B} 
            & Baseline & Random & 128 & 31.80 & 34.93 \\
            & IWI & ViT-T & 92  & 31.71 & 34.82 \\
            & CoM-PT & ViT-S & \textbf{18}  & \textbf{31.83} & \textbf{35.67} \\
            \bottomrule
        \end{tabular}
    \end{subtable}
    \hfill 
    \begin{subtable}{0.48\linewidth}
        \centering
        \setstretch{0.93}
        \caption{Complementarity of two components in inverse knowledge transfer.}
        \label{exp:ablation_com_tab}
        \begin{tabular}{l | c | c c | c c}
            \toprule
            \multirow{2}{*}{\textbf{Method}} & \multirow{2}{*}{\textbf{Epoch}} & \multicolumn{2}{c|}{\textbf{Components}} & \multicolumn{2}{c}{\textbf{ZS Performance}} \\
            & & IWI & IFD & IN-1K & COCO \\
            \midrule
            \multirow{2}{*}{Baseline} & 128 & - & - & 30.16 & 32.81 \\
            & 24  & - & - & 26.48 & 28.34 \\ 
            \midrule
            \multirow{3}{*}{CoM-PT}  & \multirow{3}{*}{24} & \checkmark & - & 28.03 & 30.44 \\
            &  & - & \checkmark & 29.29 & 32.10 \\
             &  & \checkmark & \checkmark & \textbf{30.24} & \textbf{34.15} \\
            \bottomrule
        \end{tabular}
    \end{subtable}
    \vspace{-2mm}
\end{table*}

\subsubsection{Training Epoch Allocation along the Model Chain}
\label{sec:epoch_allocation}
Next, we verify the third rule: training epoch allocation along the model chain.
By interpolating performance curves across varying epochs, we pinpoint exactly where each model matches the baseline.
As shown in Figure \ref{fig:exp_strategy}, the required training epochs for ViT-S, ViT-B, and ViT-L decrease in a roughly uniform manner, 23.45, 18.28, and 12.34 epochs, respectively. 
Notably, as we progress along the ViT-T$\rightarrow$ViT-S$\rightarrow$ViT-B$\rightarrow$ViT-L, model size grows \textbf{exponentially}, whereas the required training epochs decrease \textbf{linearly}.

\subsection{Ablation Study}

\subsubsection{Synergy of Core Designs in CoM-PT} 

\textbf{Model Chain Drives Primary Efficiency Gain.} As Table \ref{tb:com_cmp_ind} shows:
i) with inverse weight initialization alone, the independently trained ViT-S/ViT-B
needs 1.3$\times$/5.1$\times$ epochs to approach CoM-PT, losing its acceleration efficacy as model size increases; ii)  therefore, the knowledge transfer relay provided by the model chain plays the pivotal role in accelerating the training pipeline, assisted by the two components in inverse knowledge transfer applied to the current model.

\vspace{+0.25em}

\noindent \textbf{Complementarity of Two Components in Inverse Knowledge Transfer.} As emphasized in Section \ref{sec:intro}, effective inverse knowledge transfer requires knowledge reuse across both feature and parameter spaces. This is evidenced by two key phenomena: i) Table \ref{exp:ablation_com_tab} shows both components improve zero-shot performance, peaking when combined; ii) Figure \ref{exp:ablation_com_fig} reveals IWI primarily accelerates early training, whereas IFD provides sustained benefits throughout training.

\subsubsection{Pre-training Acceleration under Different Setups}
\label{exp:ablation_setups}
\begin{figure}[]
    \centering
    \vskip -0.0159 in
    \begin{subfigure}{0.21\textwidth}
        \centering
        \includegraphics[width=\textwidth]{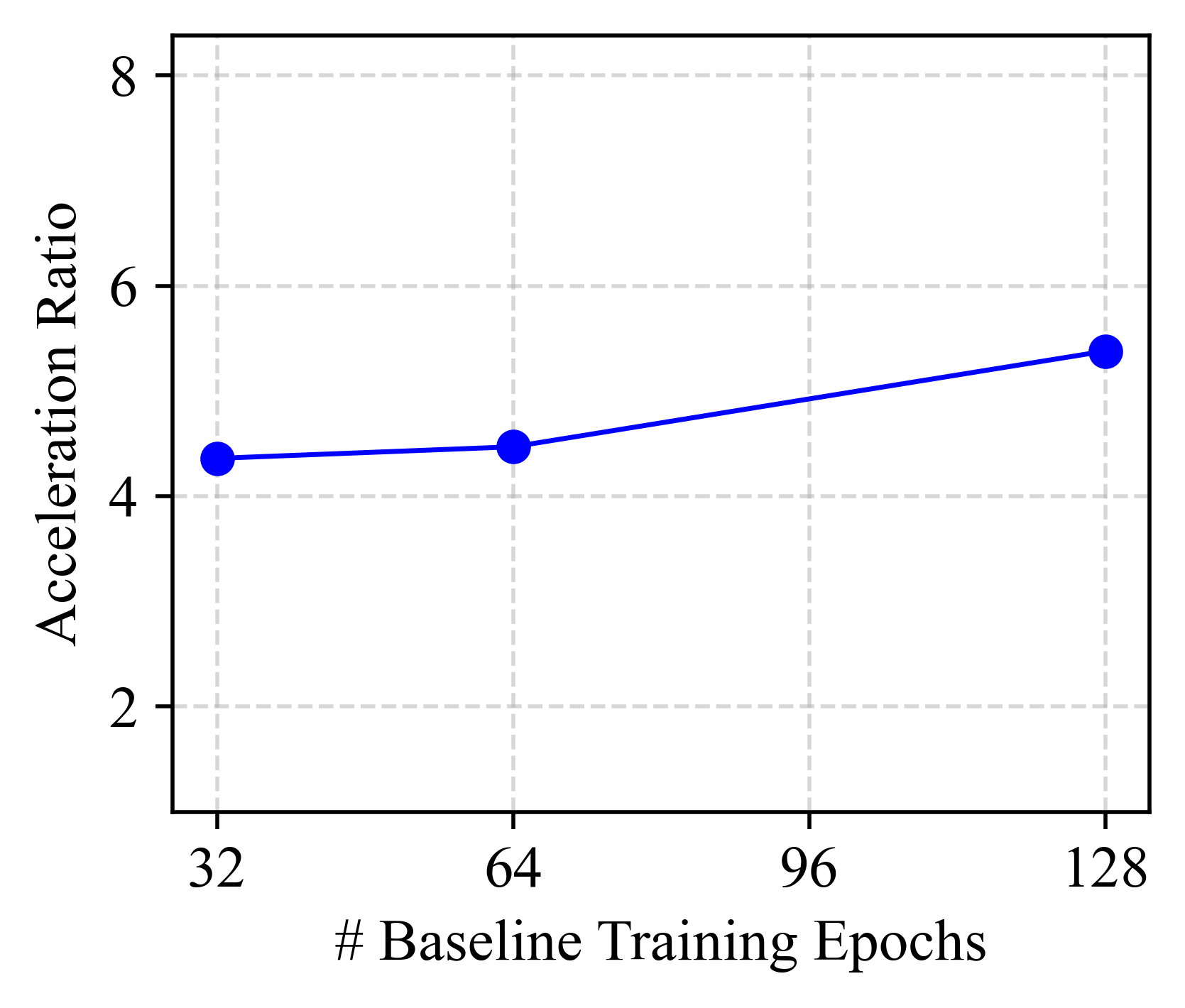}
        \vskip -0.07 in
        \caption{Number of training epochs.}        \label{fig:different_epochs}
    \end{subfigure}
    \hfill
    \begin{subfigure}{0.21\textwidth}
        \centering
        \includegraphics[width=\textwidth]{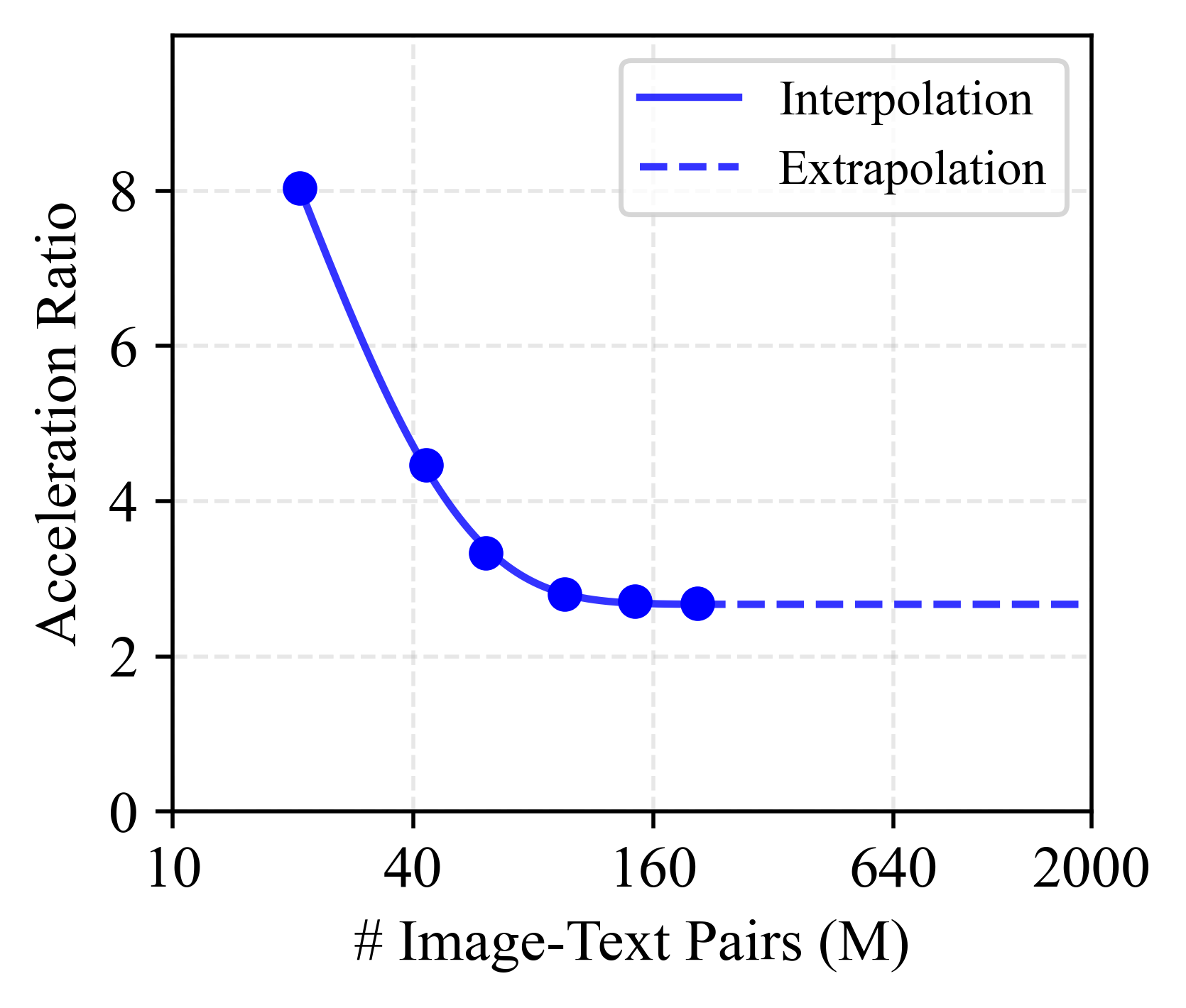}
        \vskip -0.07 in
        \caption{Number of image-text pairs.}
        \label{fig:data_scale}
    \end{subfigure}
    \vskip -0.1 in
    \caption{\textbf{Acceleration ratio variation under different training epochs and data scales.}  Experiments use the ViT family. (a) On CC3M, we vary the training epochs (32, 64, and 128). (b) We test the data scale from 22.0M to 206.8M under a fixed 64-epoch setting to isolate data scale effects. The curve fitting format follows \cite{zhai2022scaling}.}
    \label{fig:factors}
    \vskip -0.23 in

\end{figure}

\textbf{Acceleration under Various Training Epochs.} To showcase the effectiveness of CoM-PT under different training epochs, we reduce the default number of training epochs on CC3M from 128 to 64 and 32. From the results shown in Figure \ref{fig:different_epochs},  we clearly see that upon reducing the baseline training epochs, the overall acceleration stably sustains an acceleration ratio above 4$\times$. This validates that the efficacy of our method does not rely on the longer training epochs.

\vspace{+0.25em}
\noindent\textbf{Scaling-up Pre-Training Image-Text Pairs.} 
We investigate CoM-PT acceleration across different scales of pre-training image-text pairs from 22.0M (\nicefrac{1}{2} Augmented CC3M) to 206.8M (Augmented Merged-15M). With fixed baseline epochs as 64 and family size as 4, Figure \ref{fig:data_scale} shows acceleration ratio first sharply decreases and then stabilizes around 2.8$\times$ from 103.4M to 206.8M. Extrapolating to 2B scale maintains above 2.3$\times$ acceleration, showcasing the potential of CoM-PT on larger data scales. Given the limited family size, these acceleration ratios are far from the upper bound.

\vspace{-0.5mm}

\section{Conclusion}

We propose Chain-of-Models Pre-Training (CoM-PT), a novel performance-lossless training acceleration method for VFM families. This method enables expanding family size virtually for free, addressing the critical challenge of escalating pre-training costs associated with family size scaling.
Moving forward, we aim to extend our method to benefit the broader AI community, particularly in LLM pre-training.

{
    \small
    \bibliographystyle{ieeenat_fullname}
    \bibliography{main}
}
\newpage



\maketitlesupplementary

\appendix

\renewcommand{\thefigure}{\Alph{figure}}
\renewcommand{\thetable}{\Alph{table}}

\setcounter{table}{0}
\setcounter{figure}{0}

\section{Datasets} 
\label{sec:a}
\textbf{Pre-training Datasets.} CC3M \cite{sharma2018conceptual} and CC12M \cite{changpinyo2021cc12m} are two public image-text datasets, which are widely used in efficient CLIP \cite{fan2023improving, liu2023mllms, zheng2024dreamlip}. Originally, CC3M and CC12M datasets consist of around 3.3M and 12.4M image-text pairs, respectively. However, since some URLs have become invalid, we use all valid samples from each dataset, 83.0\% and 81.9\% of the original CC3M and CC12M, respectively. The details are shown in Table \ref{tb:dataset}. We incorporate long captions generated by MLLM from \cite{zheng2024dreamlip} to enhance the diversity of captions. Specifically, for each image, we select additional 15 sub-captions split from the long caption to augment the original raw data. In this way, we can expand the 2.7M image-text pairs in CC3M to approximately 44.1M image-text pairs, and the 10.2M in CC12M to approximately 162.8M image-text pairs.
We then combine the augmented CC3M and augmented CC12M to the final Merged-15M, which consists of 206.8M image-text pairs.
Under such scales, the models pre-trained by the baseline can significantly move towards the results pre-trained on extremely large datasets, such as LAION-400M \cite{schuhmann2021laion}. This approach allows us to achieve plausible acceleration ratios.

\begin{table}[h!]
\centering
\footnotesize
\caption{\textbf{CC3M, CC12M and our Merged-15M datasets used for pre-training in this paper.} Note that we only report the number of samples in the training set.}
\resizebox{0.48\textwidth}{!}{
\begin{tabular}{lccc}
\toprule
Dataset & \# Total Pairs  & \# Collected Pairs & \# Augmented Pairs\\
\midrule
CC3M & 3,318,333 & 2,753,427 & 44,054,832\\
CC12M & 12,423,374 & 10,173,631 & 162,778,096\\
Merged-15M & - & 12,927,058 & 206,832,928 \\
\bottomrule
\end{tabular}
}
\label{tb:dataset}
\end{table}


\begin{table}[h]
\centering
\caption{\textbf{Datasets in VTAB+ with abbreviations and test sizes.}}
\resizebox{0.48\textwidth}{!}{
\begin{tabular}{llr}
\toprule
Dataset & Abbr. & Test size \\
\midrule
ImageNet-1K & - & 50,000 \\
ImageNet-v2 & - & 10,000 \\
ImageNet-R & - & 30,000 \\
ImageNet Sketch & - & 50,889 \\
ObjectNet & - & 18,574 \\
ImageNet-A & - & 7,500 \\
CIFAR-10 & - & 10,000 \\
CIFAR-100 & - & 10,000 \\
MNIST & - & 10,000 \\
Oxford Flowers 102 & Flowers & 6,149 \\
Stanford Cars & Cars & 8,041 \\
SVHN & - & 26,032 \\
Facial Emotion Recognition 2013 & FER2013 & 7,178 \\
RenderedSST2 & RSST2 & 1,821 \\
Oxford-IIIT Pets & Pets & 3,669 \\
Caltech-101 & - & 6,085 \\
Pascal VOC 2007 Classification & VOC2007 & 14,976 \\
SUN397 & - & 108,754 \\
FGVC Aircraft & Aircraft & 3,333 \\
Country211 & - & 21,100 \\
Describable Textures & DTD & 1,880 \\
GTSRB & - & 12,630 \\
STL10 & - & 8,000 \\
Diabetic Retinopathy & Retino & 42,670 \\
EuroSAT & - & 5,400 \\
RESISC45 & - & 6,300 \\
PatchCamelyon & PCAM & 32,768 \\
CLEVR Counts & - & 15,000 \\
CLEVR Object Distance & CLEVR Dist & 15,000 \\
DSPRITES Orientation & DSPRITES Orient & 73,728 \\
DSPRITES Position & DSPRITES pos & 73,728 \\
SmallNORB Elevation & SmallNORB Elv & 12,150 \\
SmallNORB Azimuth & SmallNORB Azim & 12,150 \\
DMLAB & - & 22,735 \\
KITTI closest vehicle distance & KITTI Dist & 711 \\
\bottomrule
\end{tabular}
}
\vspace{-0.2in}
\label{tab:datasets}
\end{table}

\noindent\textbf{Zero-shot Classification and Retrieval Datasets.} As we mentioned in the main paper, we follow the standard evaluation protocols \cite{ cherti2023reproducible} for all datasets: ImageNet-1K \cite{deng2009imagenet} for zero-shot classification,  COCO \cite{lin2014microsoft} for zero-shot image-text retrieval, and VTAB+ \cite{schuhmann2022laion} for testing zero-shot transfer capabilities. 
Specifically, VTAB+ is one of the largest zero-shot benchmarks so far, which consists of 35 datasets, as shown in Table \ref{tab:datasets}. The evaluation is conducted on the official CLIP benchmark \cite{cherti_2025_15403103}.

\begin{table*}[]
    \centering
    \footnotesize
    \caption{\textbf{Detailed settings of baseline pre-training and chain-of-models pre-training on CC3M and Merged-15M datasets.}}
        \centering
        \footnotesize
        \vspace{-0.05in}
        \subfloat[Baseline training strategy.]{
            \begin{tabular}{l|c}
            \toprule
            \textbf{Configuration}  &  \textbf{Setting} \\ \midrule
             Dataset & CC3M $|$ Merged-15M \\ \midrule
             Batch Size &  1024 \\
             Optimizer & AdamW \\
             Optimizer Hyper-Parameters & $\beta_1, \beta_2$=(0.9, 0.98), $\epsilon$=1e-8 \\
             Learning Rate Schedule & Cosine Decay \\
             Initial Learning Rate & 1e-3 \\ 
             Weight Decay & 0.1 \\
             Training Epochs & 128$|$64 \\
             Warmup Iterations & 8000 \\ 
             Precision & AMP \\
             \bottomrule
            \end{tabular}
        }
        \vspace{-0.05in}
    \hfill
        \centering
        \footnotesize
        \vspace{-0.05in}
        \subfloat[CoM-PT training strategy.]{
            \begin{tabular}{l|c}
            \toprule
            \textbf{Configuration}  &  \textbf{Setting} \\ \midrule
             Dataset & CC3M $|$ Merged-15M  \\ \midrule
             Batch Size &  1024 \\
             Optimizer & AdamW \\
             Optimizer Hyper-Parameters & $\beta_1, \beta_2$=(0.9, 0.98), $\epsilon$=1e-8 \\
             Learning Rate Schedule & Cosine Decay \\
             Initial Learning Rate & 1e-3 \\ 
             Weight Decay & 0.1 \\
             Training Epochs & \textit{In Table \ref{tb:training_epochs}} \\
             Warmup Iterations & 1000 \\ 
             Precision & AMP \\
             \bottomrule
            \end{tabular}
        }
\label{tb:pre-training settings}
\end{table*}

\begin{table}[htbp]
\centering
\small
\renewcommand{\arraystretch}{1.2}
\vspace{-0.05in}
\caption{\textbf{Training epochs for different models in main experiments.}}
\vspace{-0.05in}
\resizebox{0.45\textwidth}{!}{%
\begin{tabular}{l|ccccc|cccc}
\toprule
\multirow{2}{*}{\textbf{Dataset}} & \multicolumn{5}{c|}{\textbf{ViT Family}} & \multicolumn{4}{c}{\textbf{Swin Family}} \\
& \rotatebox{90}{ViT-T/16} & \rotatebox{90}{ViT-S/16} & \rotatebox{90}{ViT-M/16} & \rotatebox{90}{ViT-B/16} & \rotatebox{90}{ViT-L/16} & \rotatebox{90}{Swin-T} & \rotatebox{90}{Swin-S} & \rotatebox{90}{Swin-B} & \rotatebox{90}{Swin-L} \\
\midrule
CC3M & 128 & 24 & -- & 18 & 15 & 128 & 24 & 16 & 12 \\
Merged-15M & -- & 64 & 21 & 17 & 15 & -- & -- & -- & -- \\
\bottomrule
\end{tabular}
}
\vspace{-0.2in}
\label{tb:training_epochs}
\end{table}

\vspace{+0.2em}

\noindent\textbf{Datasets for Open-vocabulary Semantic Segmentation Tasks.} We follow the training and evaluation protocol of SAN.
i) We train the model on the training set of COCO-Stuff \cite{caesar2018coco}, which contains 118K images with 171 annotated classes. ii) We test the model on ADE-847 \cite{zhou2017scene}, ADE-150 \cite{zhou2017scene}, PC-459 \cite{mottaghi2014role}, PC-59 \cite{mottaghi2014role}, and VOC-20 \cite{everingham2011pascal}. Specifically, ADE-150 and ADE-847 have the same 2K validation images, but have different numbers of categories, 150 and 847 annotated classes, respectively. Similarly, PC-59 and PC-459 have 5K validation images, but have 59 and 459 classes, respectively. And VOC-20 contains 1449 images with 20 classes.

\vspace{+0.2em}

\noindent\textbf{Datasets for Vision-Language Tasks.} The training process of LLaVA-1.5 \cite{liu2024improved} consists of two main stages: quick feature alignment training and visual instruction fine-tuning. For feature alignment training, we utilize the LLaVA-Pretrain LCS-558K dataset \cite{liu2024improved}, comprising 558K image-text pairs filtered from the original LAION~\cite{schuhmann2022laion}, CC12M~\cite{changpinyo2021cc12m}, and SBU~\cite{ordonez2011im2text} datasets, with BLIP-generated captions~\cite{li2023blip}. Subsequently, visual instruction tuning is performed using the LLaVA-Instruct-150K dataset \cite{liu2024improved}, which contains 150K GPT-4-generated multimodal instruction-following samples \cite{achiam2023gpt}. We assess the model's performance across several downstream tasks: TextVQA~\cite{singh2019towards}, which evaluates the model's ability to read and reason about text within images; ScienceQA~\cite{lu2022learn}, a benchmark with multimodal multiple-choice science questions testing the model's scientific reasoning capabilities; POPE~\cite{li2023evaluating}, designed to evaluate object hallucination in vision-language models; and VQAv2~\cite{goyal2017making}, a dataset containing open-ended questions about images, requiring models to integrate visual understanding, language processing, and commonsense knowledge.


\section{Experimental Setups}

\label{sec:b}
\subsection{Contrastive Language Image Pre-training }

\textbf{Compute Infrastructure.} All experiments are conducted on a single node equipped with $8 \times$ NVIDIA A100 (80GB) GPUs.

\vspace{+0.2em}

\noindent\textbf{Implementation Codebase.} Our chain-of-models pre-training is implemented using the OpenCLIP framework~\cite{cherti2023reproducible}.

\vspace{+0.2em}

\noindent\textbf{Hyper-parameter Settings.} The training loss for our experiments is defined in Equation \ref{eq1} and Equation \ref{eq4} of the main paper, which only has one hyper-parameter $\alpha$. Based on the results in the Section \ref{sec:ifd}, we choose $\alpha=500$ (corresponding to $r=0.1$ in Figure \ref{exp:abl_ifd}) for all experiments.

\vspace{+0.2em}

\noindent\textbf{Training Strategy.} Following the standard training protocol of OpenCLIP~\cite{cherti2023reproducible}, automatic mixed precision (AMP) is applied as default. Besides that, to ensure fair acceleration ratios, all experiments on CC3M and Merged-15M datasets are typically conducted with the same setting, including data processing pipeline, optimizer, initial learning rate, batch size, \textit{etc}. Details are shown in Table \ref{tb:pre-training settings}.

\vspace{+0.2em}

\noindent\textbf{Training Arrangement of CoM-PT.} Derived from the ablation study regarding training arrangement in Section \ref{sec:epoch_allocation} of the main paper, we have summarized: \textit{"training epochs allocation along the model chain decreases linearly as model size increases exponentially."} Following this principle, we arrange the training epochs for each model as shown in Table  \ref{tb:training_epochs}. Notably, compared to the interpolated minimum epochs on CC3M (23.45, 18.28, and 12.34 for ViT-S/16, ViT-B/16, and ViT-L/16, respectively), we allocate additional epochs to the largest model in practice to obtain superior performance in the main experiments.

\noindent\subsection{Side Fine-tuning on SAN}
\begin{table*}[]
    \centering
    \footnotesize
    \caption{\textbf{Detailed settings of fine-tuning on COCO-Stuff for open-vocabulary semantic segmentation tasks.}}
    \begin{minipage}{0.49\textwidth}
        \centering
        \subfloat[SAN with ViT-B/16 backbone.]{
            \begin{tabular}{l|c}
            \toprule
            \textbf{Configuration}  &  \textbf{Setting} \\ \midrule
             Segmentation Framework & SAN  \\ 
             Backbone & ViT-B/16  \\   
             Feature Fusion Blocks & 1-9  \\   
             Mask Recognition Blocks & 10-12 \\ \midrule
             Input Resolution & 640 \\
             Batch Size &  32 \\
             Optimizer & AdamW \\
             Optimizer Hyper-Parameters & $\beta_1, \beta_2$=(0.9, 0.98), $\epsilon$=1e-8 \\
             Learning Rate Schedule & Poly Decay \\
             Initial Learning Rate & 1e-3 \\ 
             Learning Rate Decay & 0.9 \\
             Weight Decay & 1e-4 \\
             Training Iterations & 60K \\
             Random Resize Crop & \checkmark \\ 
             Precision & AMP \\
             \bottomrule
            \end{tabular}
        }
    \end{minipage}
    \hfill
    \begin{minipage}{0.49\textwidth}
        \centering
        \subfloat[SAN with ViT-L/16 backbone.]{
            \begin{tabular}{l|c}
            \toprule
            \textbf{Configuration}  &  \textbf{Setting} \\ \midrule
             Segmentation Framework & SAN  \\ 
             Backbone & ViT-L/16  \\   
             Feature Fusion Blocks & 1-18  \\   
             Mask Recognition Blocks & 19-24 \\ \midrule
             Input Resolution & 640 \\
             Batch Size &  32 \\
             Optimizer & AdamW \\
             Optimizer Hyper-Parameters & $\beta_1, \beta_2$=(0.9, 0.98), $\epsilon$=1e-8 \\
             Learning Rate Schedule & Poly Decay \\
             Initial Learning Rate & 1e-3 \\ 
             Learning Rate Decay & 0.9 \\
             Weight Decay & 1e-4 \\
             Training Iterations & 60K \\
             Random Resize Crop & \checkmark \\ 
             Precision & AMP \\
             \bottomrule
            \end{tabular}
        }
    \end{minipage}
\label{tb:semantic_segmentation_settings}
\end{table*}

\begin{table*}[]
    \centering
    \footnotesize
    \caption{\textbf{Detailed settings of feature alignment training and LoRA-based visual instruction fine-tuning.}}
    \begin{minipage}{0.49\textwidth}
        \centering
        \footnotesize
        \vspace{-0.05in}
        \subfloat[Feature alignment training. \label{tab:feature-alignment}]{
            \begin{tabular}{l|c}
            \toprule
            \textbf{Configuration}  &  \textbf{Setting} \\ \midrule
             Multi-Modality Large Language Model & LLaVA-1.5-7B \\
             Vision Encoder & ViT-B/16 | ViT-L/16 \\
             Feature Alignment Projector & mlp2x\_gelu \\ \midrule
             Model Max Length & 2048 \\
             LoRA Rank & -- \\
             LoRA Alpha & -- \\ 
             Per Device Batch Size  & 48 \\ 
             Gradient Accumulation Steps & 1 \\
             Optimizer & AdamW \\
             Feature Alignment Projector LR & 2e-3 \\
             LoRA LR &  -- \\
             Learning Rate Schedule & Cosine Decay \\
             Warmup Ratio  & 0.03 \\
             Training Epochs  & 1 \\
             BF16 & \checkmark \\
             \bottomrule
            \end{tabular}
        }
        \vspace{-0.05in}
    \end{minipage}
    \hfill
    \begin{minipage}{0.49\textwidth}
        \centering
        \footnotesize
        \vspace{-0.05in}
        \subfloat[LoRA-based visual instruction fine-tuning. \label{tab:visual-tuning}]{
            \begin{tabular}{l|c}
            \toprule
            \textbf{Configuration}  &  \textbf{Setting} \\ \midrule
             Multi-Modality Large Language Model & LLaVA-1.5-7B \\
             Vision Encoder & ViT-B/16 | ViT-L/16 \\
             Feature Alignment Projector & mlp2x\_gelu \\  \midrule
             Model Max Length & 2048 \\ 
             LoRA Rank & 128 \\
             LoRA Alpha & 256 \\ 
             Per Device Batch Size  & 16 \\ 
             Gradient Accumulation Steps & 1 \\
             Optimizer & AdamW \\
             Feature Alignment Projector LR & 2e-5 \\
             Lora LR &  2e-4 \\
             Learning Rate Schedule & Cosine Decay \\
             Warmup Ratio  & 0.03 \\
             Training Epochs  & 1 \\
             BF16 & \checkmark \\
             \bottomrule
            \end{tabular}
        }
        \vspace{-0.05in}
    \end{minipage}
\end{table*}

\vspace{+0.2em}

\noindent\textbf{Compute Infrastructure.} The experiments are conducted on a single node equipped with $8 \times$ NVIDIA A100 (80GB) GPUs.

\vspace{+0.2em}

\noindent\textbf{Implementation Codebase.} We conduct experiments using the official repository of SAN~\cite{xu2023san}.

\vspace{+0.2em}

\noindent\textbf{Selection of Segmentation Frameworks and Backbones.} We choose SAN as the segmentation framework based on the pre-trained ViT-B/16 and ViT-L/16 backbones.

\vspace{+0.2em}

\noindent\textbf{Training Strategy.} We directly use the official fine-tuning strategy of SAN. Details are shown in Table \ref{tb:semantic_segmentation_settings}.

\subsection{Visual Instruction Fine-tuning on LLaVA}

\noindent\textbf{Compute Infrastructure.} The experiments are conducted on a single node equipped with $8 \times$ NVIDIA A100 (80GB) GPUs.

\vspace{+0.2em}

\noindent\textbf{Implementation Codebase.} We conduct the experiments using the official repository of LLaVA~\cite{liu2023llava,liu2024improved}.

\vspace{+0.2em}

\noindent\textbf{Selection of MLLM and Backbones.} We select LLaVA-1.5-7B~\cite{liu2024improved} as the base MLLM, equipped with either ViT-B/16 or ViT-L/16 as the image encoder.

\vspace{+0.2em}

\noindent\textbf{Training Strategy.} The training consists of two stages. i) In the first stage, we align the features of the image encoder and the LLM by fully fine-tuning a lightweight two-layer MLP vision-language connector. The experimental setup is detailed in Table~\ref{tab:feature-alignment}. ii) In the second stage, we perform efficient visual instruction fine-tuning on the Vicuna-7B-v1.5~\cite{zheng2023judging} using a LoRA-based \cite{hu2022lora} approach to enable its capability in handling vision-related tasks. The corresponding configuration is provided in Table~\ref{tab:visual-tuning}.

\begin{table*}[]
\centering
\renewcommand{\arraystretch}{1.2}
\caption{\textbf{Architecture specifications for VFM families.} \textit{Forward MACs} and \textit{Training MACs} represent the computational complexity of the model for a sample (an image and its corresponding 4 text descriptions) in our experiment. 
}
\resizebox{0.7\textwidth}{!}{%
\begin{tabular}{ll|ccc|ccc|ccc}
\toprule
Model Family & Model   & \multicolumn{3}{c|}{Image Encoder} & \multicolumn{3}{c|}{Text Encoder} & \multicolumn{3}{c}{Overall} \\
             &         & \rotatebox{90}{Width} & \rotatebox{90}{Depth} & \rotatebox{90}{Param (M)} & \rotatebox{90}{Width} & \rotatebox{90}{Depth} & \rotatebox{90}{Param (M)} & \rotatebox{90}{Param (M)} & \rotatebox{90}{Forward MACs (G)} & \rotatebox{90}{Training MACs (G)} \\
\midrule
\multirow{7}{*}{ViT Family}  & ViT-T/16 & 192   & 12    & 5.62      & 256   & 12    & 9.48      & 15.10 & 5.82           & 17.57         \\
             & ViT-C/16 & 256   & 12    & 9.86      & 320   & 12    & 14.80     & 24.66 & 9.23           & 27.87         \\
             & ViT-S/16 & 384   & 12    & 21.81     & 384   & 12    & 21.29     & 43.10 & 14.44           & 43.65        \\
             & ViT-M/16 & 512   & 12    & 38.59     & 448   & 12    & 28.97     & 67.56 & 21.88           & 66.17        \\
             & ViT-B/16 & 768   & 12    & 86.19     & 512   & 12    & 37.83     & 124.02 & 35.09          & 106.27      \\
             & ViT-XB/16 & 1024  & 12    & 171.36    & 512   & 12    & 37.83     & 209.19 & 48.76          & 148.05      \\
             & ViT-L/16 & 1024  & 24    & 304.09    & 768   & 12    & 85.05     & 389.14 & 100.92          & 306.11      \\ \midrule
\multirow{4}{*}{Swin Family} & Swin-T   & 768   & 12    & 27.91     & 256   & 12    & 9.48      & 37.39 & 8.76           & 26.60       \\
             & Swin-S   & 768   & 24    & 49.23     & 384   & 12    & 21.29     & 70.52 & 18.40           & 55.82       \\
             & Swin-B   & 1024  & 24    & 87.27     & 512   & 12    & 37.83     & 125.10 & 32.71          & 98.84       \\
             & Swin-L   & 1536  & 24    & 195.78    & 768   & 12    & 85.05     & 280.83 & 73.48          & 222.94      \\
\bottomrule
\end{tabular}
}
\label{tb:architecture}
\vspace{-1mm}
\end{table*}

\section{Architectural Specifications and Training Complexity of VFM Families}
\label{sec:c}

\subsection{Model Architectures}
Recall that we chose the standard ViT family and the Swin family in the main experiments. Detailed information about these two model families is provided in Table \ref{tb:architecture}. Note that the forward MACs are calculated based on 224$\times$224 resolution images and  77 text tokens. Yet, with the inclusion of additional sub-captions, the forward MACs from the text encoder become $4\times$ than only using the original caption.

\subsection{Calculation of Training Complexity}
As we stated in the main paper, the training complexity $C_t$ comprises the forward complexity $C_f$, the backward complexity $C_b$, and the parameter update complexity $C_u$ concerning both models and optimizers. The training complexity $C_t$ is formulated as:
\begin{equation}
    C_t = C_b + C_f + C_u.
\end{equation}
Considering multiply–accumulate operations (MACs), the backward complexity is approximately $2\times$ larger than the forward complexity for each layer, except the first layer. For a network with $p$ layers, we can derive the following equation:
\begin{equation}
    C_b = \sum_{i=1}^p C_b^i = 2\sum_{i=1}^p C_f^i - C_f^1 = 2C_f - C_f^1,
\end{equation}
where $C_f^i$ and $C_b^i$ denote the forward complexity and the backward complexity of layer $i$, respectively. Since we select AdamW \cite{loshchilov2017decoupled} as the optimizer, the MACs of the parameter update $C_u$ are $3\times$ the number of parameters. In this way, we can calculate the accurate $C_t$ for each model. 

For CoM-PT, the total training complexity is computed as a summation of student's full training complexity and teacher's forward complexity for inverse knowledge transfer.

\begin{figure*}
    \centering
    \includegraphics[width=0.78\textwidth]{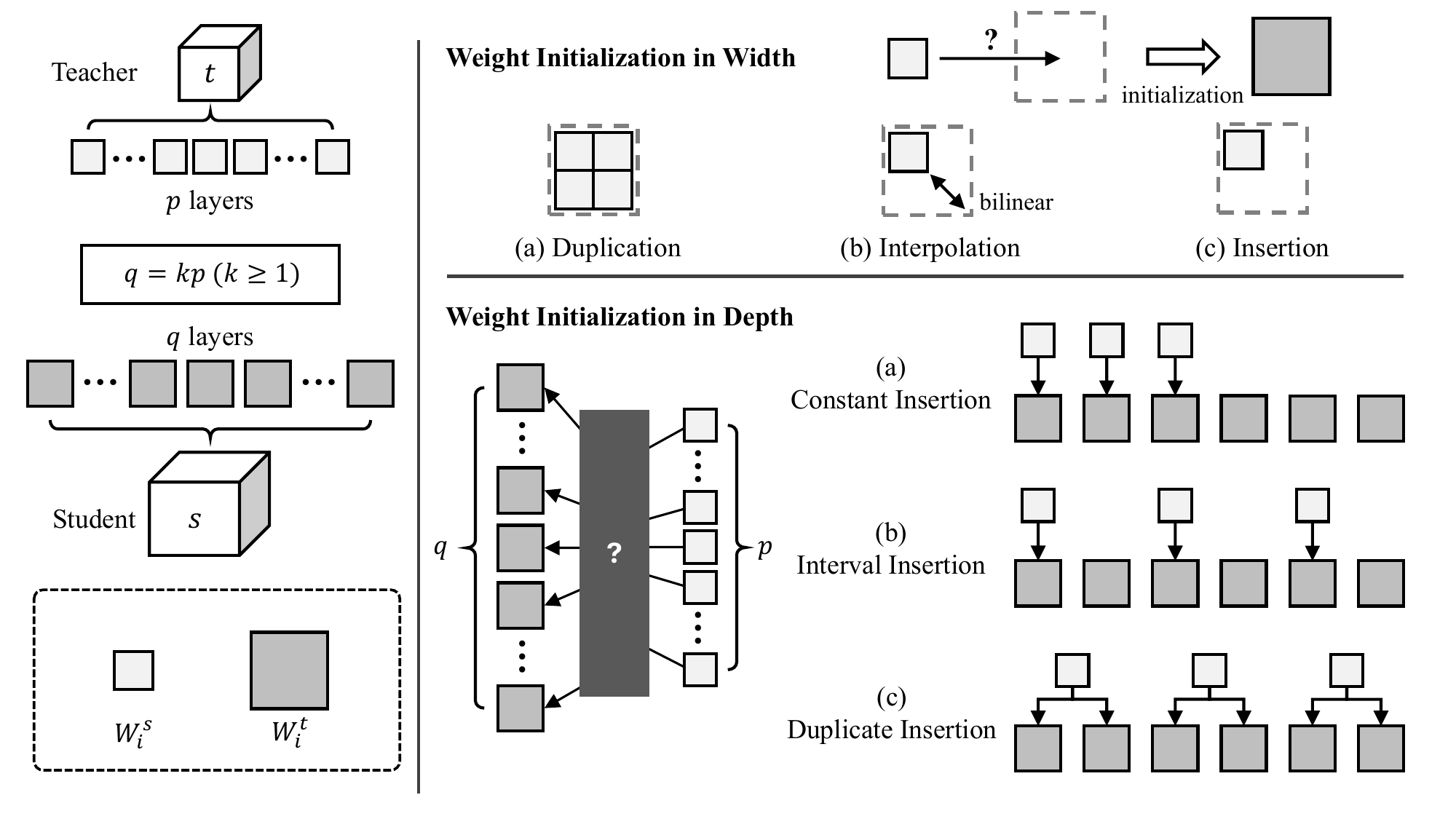}
    \caption{\textbf{Detailed illustrations of weight initialization in width and depth.} First, (a)-(c) illustrates different weight initialization approaches for width differences. Duplication (a) involves copying the teacher's weights multiple times and directly assigning these copies to the student. Interpolation (b) conducts bilinear operations to expand the teacher's weights to fit the shape of the student. Insertion (c) involves directly assigning the teacher's weights to the student, with all remaining weights initialized randomly. Second, (d)-(f) show the different weight initialization methods for depth differences. (d) only initializes the first $p$ layers across the total $q$ layers, while the interval insertion (e) skips several layers by calculating $q/p$. As a combination, the duplicate insertion (f) initializes the skipped layers by duplicating the weights of the preceding layer.}
    \label{fig:wi-method}
\end{figure*}

\section{More Ablation Studies}

\label{sec:more ablation}

\subsection{Inverse Weight Initialization}
In this part, we address two key questions regarding our implementation of inverse weight initialization.
\begin{itemize}
    \item[i)]  What kind of simple approach yields the best performance?  
    \item[ii)] Why is applying sophisticated implementations less important in our inverse knowledge transfer?
\end{itemize}

\subsubsection{Exploration of Effective Simple Approaches}
We explore several simple approaches for inverse weight initialization. As illustrated in Figure \ref{fig:wi-method}, weight initialization for width is shown from (a) to (c), and weight initialization for depth is shown from (d) to (f). Then, we conduct experiments to compare the effectiveness of these approaches.

\begin{figure} 
    \vspace{-3.0mm}
    \centering
    \includegraphics[width=0.43\textwidth]{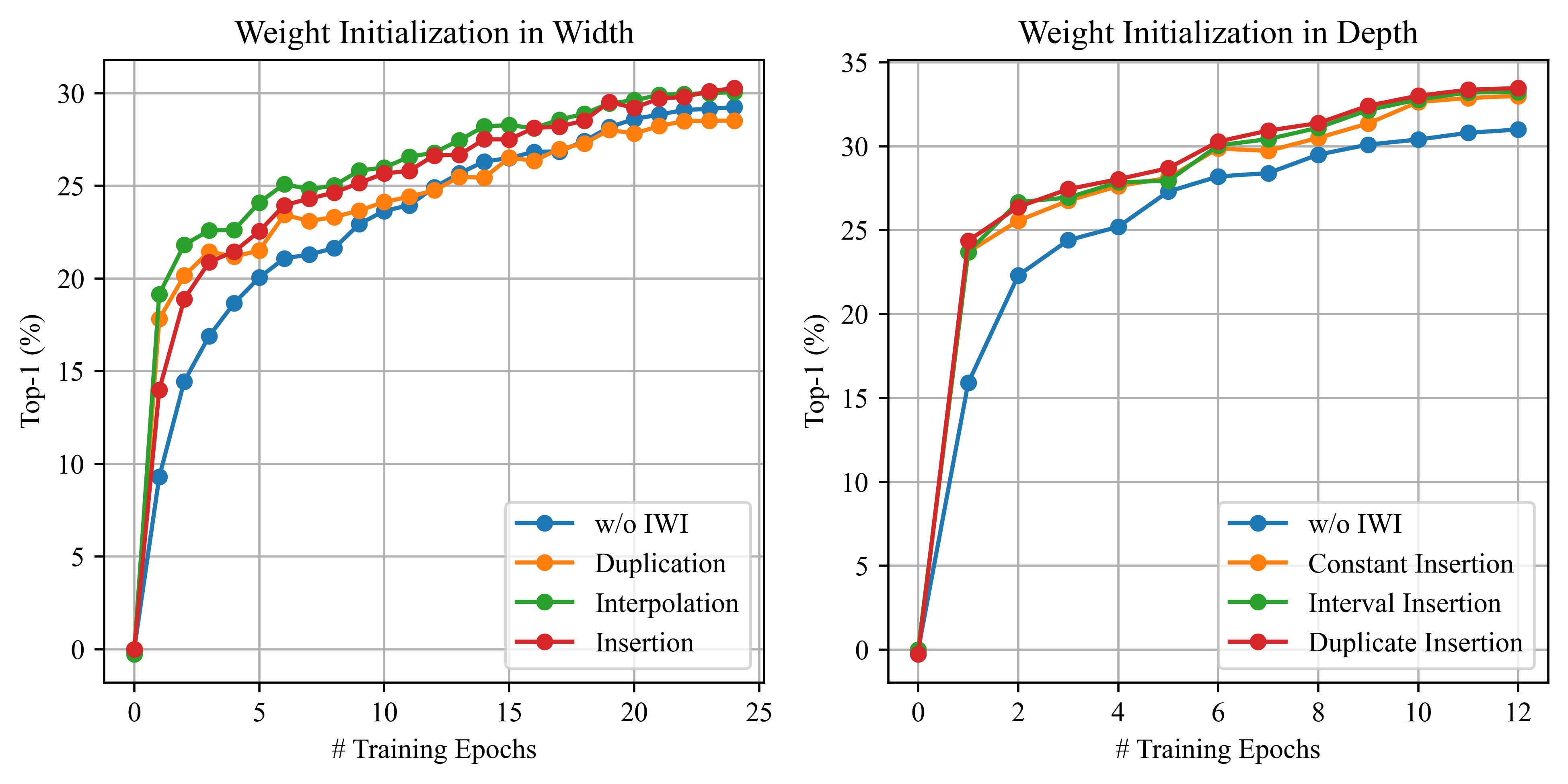}
    \caption{\textbf{Ablation study on various simple weight initialization approaches.} For width initialization, experiments are conducted on the ViT-T/16$\rightarrow$ViT-S/16 sub-chain. Progressively, with the best simple width initialization method, we explore depth initialization on the ViT-B/16$\rightarrow$ViT-L/16 sub-chain. \textit{Inverse feature distillation is applied as default. Details for each method are in Figure \ref{fig:wi-method}.}}
    \label{fig:wi}
    \vspace{-1mm}
\end{figure}

To systematically evaluate these approaches, we conduct a two-stage analysis in Figure \ref{fig:wi}. First, we examine width initialization methods within the ViT-T/16$\rightarrow$ViT-S/16 sub-chain. Our findings reveal: i) weight duplication initially accelerates convergence but ultimately harms final performance; ii) simple insertion and linear interpolation improve convergence speed and final performance, with linear interpolation excelling in convergence acceleration while insertion achieves better final results. Subsequently, we study depth initialization based on simple insertion in width. We find that, compared to constant and interval insertion in depth, duplicate insertion delivers the best performance. 

Based on this systematic evaluation, we adopt simple insertion for width differences and duplicate insertion for depth differences as our default implementation strategy.

\begin{table}
    \vspace{-3.8mm}
    \centering
    \scriptsize
    \caption{\textbf{Our simple approaches \textit{vs.} previous sophisticated designs.} Experiments are conducted using the ViT-T/16$\rightarrow$ViT-S/16 sub-chain, trained on the CC3M dataset.}
    \begin{tabular}{l|cc}
        \toprule
        \multirow{2}{*}{\textbf{Method}} & \multicolumn{2}{c}{\textbf{ZS Performance}} \\
        & ImageNet-1K & COCO \\
        \midrule
        Baseline & 26.48   & 28.34 \\ \midrule
        +IFD & 29.29 & 32.10 \\
        +IFD+IWI  & 30.24  & \textbf{34.15}  \\
        \midrule
        +IFD+Net2Net \cite{chen2015net2net} & 30.07   & 33.63 \\
        +IFD+NetExpand \cite{ding2023network} & \textbf{30.36}  & 34.13 \\
        \bottomrule
    \end{tabular}
\label{exp:ablation_other}
\vspace{-4mm}
\end{table}

\begin{figure}
    \centering
    \includegraphics[width=0.35\textwidth]{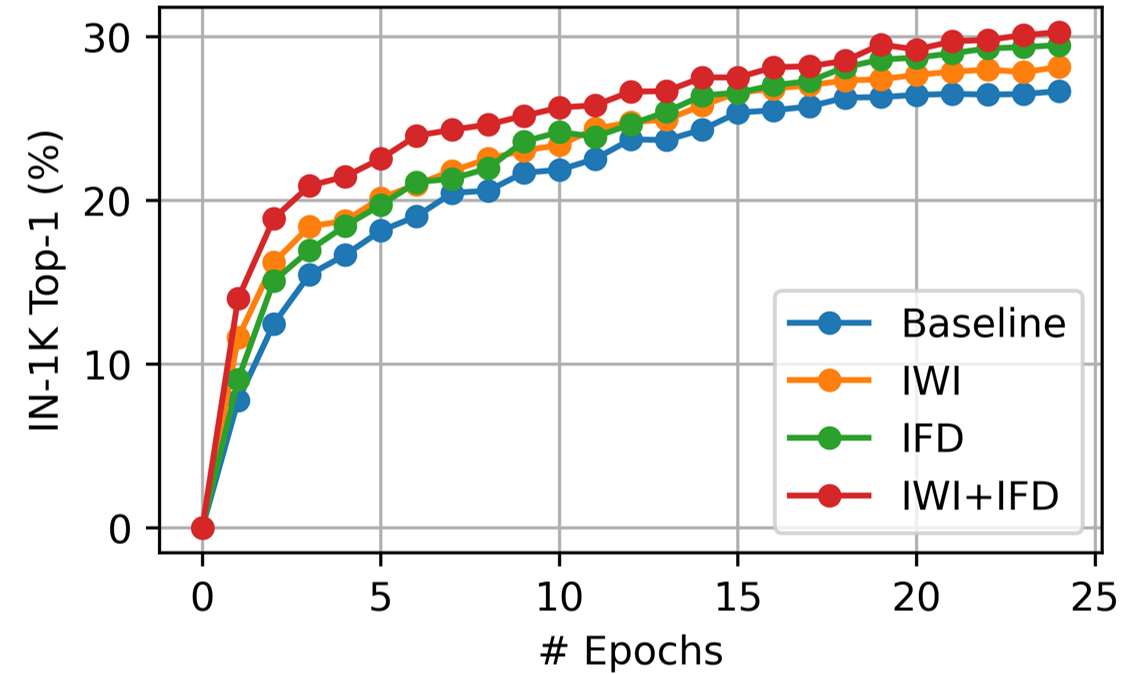}
    \vspace{-3mm}
    \caption{\textbf{Validation curves of different choices in Table \ref{exp:ablation_com_tab}} of the main paper.}
    \vspace{-5mm}
    \label{exp:ablation_com_fig}
\end{figure}

\begin{figure*}
    \centering
    \subfloat[Performance of ViT-S/16 under different $r$.]{
    \includegraphics[width=0.50\textwidth]{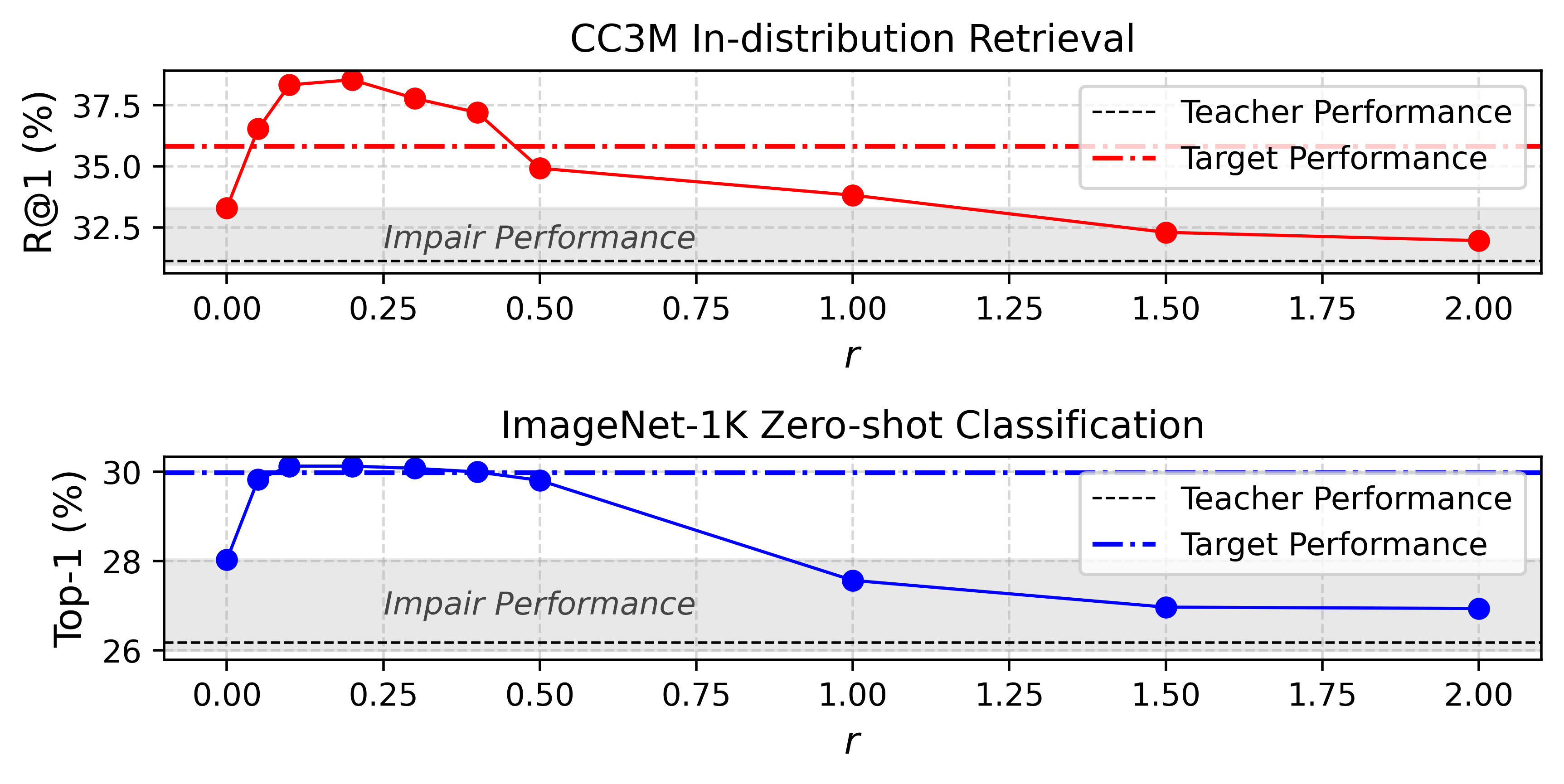}
    \label{exp:abl_ifd_a}
    }
    \subfloat[Loss curves $r=0.1$.]{
    \includegraphics[width=0.28\textwidth]{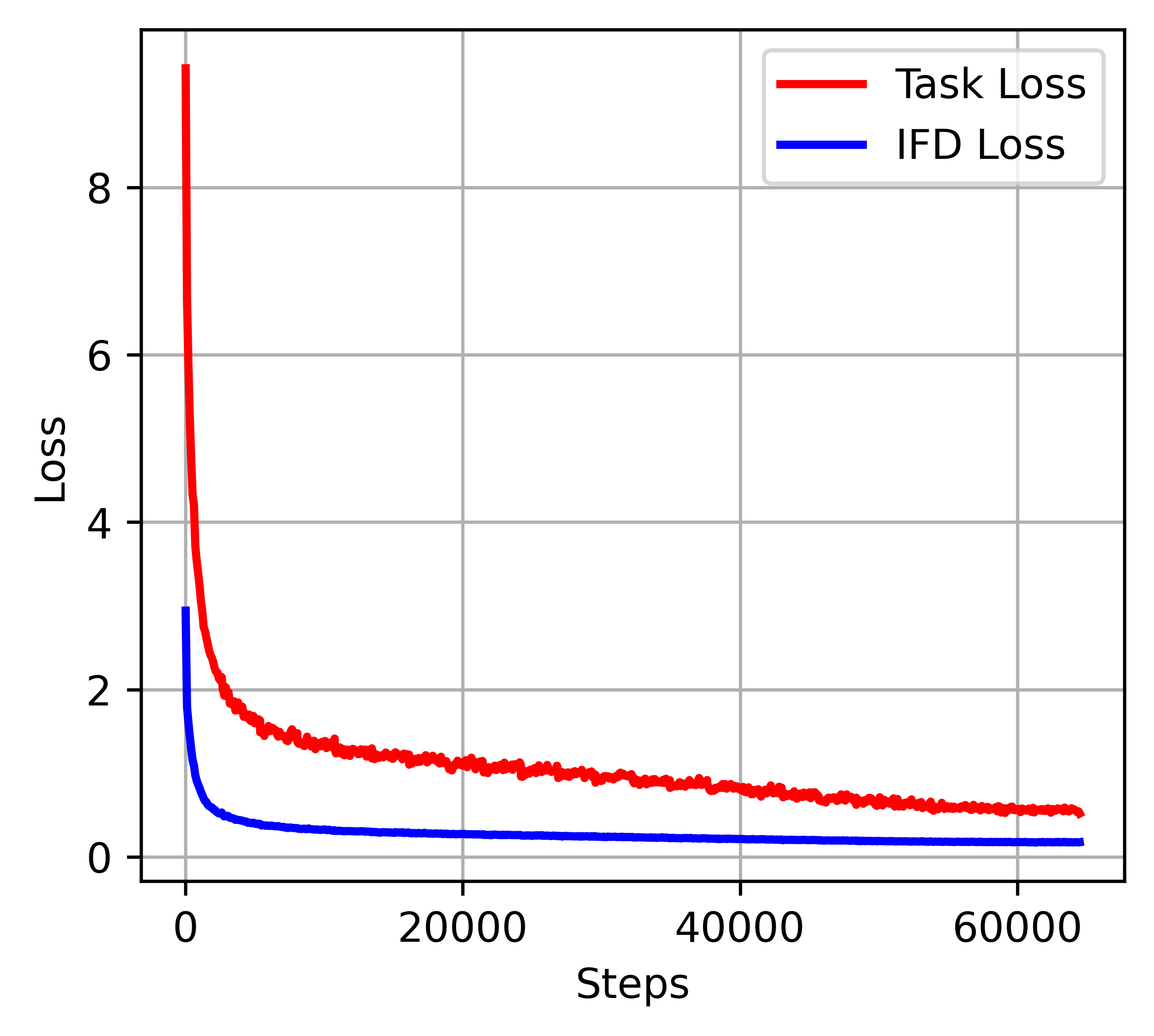}
    \label{exp:abl_ifd_b}
    }
    \vskip -0.05 in
    \caption{\textbf{Ablation study on inverse feature distillation magnitude.} For better illustration, we define $r$ as the ratio between the magnitude of $L_{IFD}$ and $L_{task}$.
    The experiment is conducted on ViT-T/16$\rightarrow$ViT-S/16 on the CC3M dataset. We show how IFD enables ViT-S/16 to achieve target performance (128-epoch individual pre-training) with only 24 epochs. The teacher model is ViT-T/16 trained individually for 128 epochs. 
    \textit{Inverse weight initialization is applied as default.}}
    \label{exp:abl_ifd}
\end{figure*}

\begin{table*}[h]
\centering
\renewcommand{\arraystretch}{1.3}
\caption{
\textbf{Zero-shot Top-1 accuracy (\%) across the 35 datasets of the VTAB+ benchmark.} Detailed results corresponding to the VTAB+ averages presented in Table \ref{tab:main} of the main paper.
}
\resizebox{\textwidth}{!}{%
\begin{tabular}{l|cccccccccccccc|ccccccccccc}
\toprule
PT Dataset & \multicolumn{14}{c|}{CC3M} & \multicolumn{7}{c}{Merged-15M} \\
\midrule
Models & ViT-T/16 & \multicolumn{2}{c}{ViT-S/16} & \multicolumn{2}{c}{ViT-B/16} & \multicolumn{2}{c}{ViT-L/16} & Swin-T & \multicolumn{2}{c}{Swin-S} & \multicolumn{2}{c}{Swin-B} & \multicolumn{2}{c|}{Swin-L} & ViT-S/16 & \multicolumn{2}{c}{ViT-M/16} & \multicolumn{2}{c}{ViT-B/16} & \multicolumn{2}{c}{ViT-L/16} \\
\midrule
PT Method & \rotatebox{90}{baseline} & \rotatebox{90}{baseline} & \rotatebox{90}{CoM-PT} & \rotatebox{90}{baseline} & \rotatebox{90}{CoM-PT} & \rotatebox{90}{baseline} & \rotatebox{90}{CoM-PT} & \rotatebox{90}{baseline} & \rotatebox{90}{baseline} & \rotatebox{90}{CoM-PT} & \rotatebox{90}{baseline} & \rotatebox{90}{CoM-PT} & \rotatebox{90}{baseline} & \rotatebox{90}{CoM-PT} & \rotatebox{90}{baseline} & \rotatebox{90}{baseline} & \rotatebox{90}{CoM-PT} & \rotatebox{90}{baseline} & \rotatebox{90}{CoM-PT} & \rotatebox{90}{baseline} & \rotatebox{90}{CoM-PT} \\
\midrule
ImageNet-1K & 26.16 & 30.16 & 30.24 & 31.80 & 31.83 & 33.77 & 34.27 & 33.84 & 35.88 & 36.19 & 36.13 & 36.71 & 36.77 & 37.06 & 43.97 & 45.90 & 45.85 & 47.15 & 47.39 & 48.80 & 49.47 \\
ImageNet-v2 & 22.85 & 26.47 & 25.91 & 27.42 & 28.00 & 29.39 & 29.94 & 29.00 & 30.99 & 31.31 & 31.54 & 31.59 & 31.83 & 32.30 & 38.24 & 39.52 & 39.85 & 41.35 & 40.79 & 43.06 & 42.84 \\
ImageNet-R & 37.15 & 42.36 & 40.77 & 44.09 & 43.76 & 47.60 & 47.94 & 47.35 & 50.70 & 50.72 & 52.16 & 52.04 & 51.85 & 52.96 & 63.28 & 66.06 & 65.61 & 69.16 & 67.70 & 71.65 & 71.02 \\
ImageNet Sketch & 16.93 & 21.24 & 20.53 & 22.81 & 21.32 & 24.54 & 24.76 & 24.92 & 26.25 & 26.80 & 27.66 & 27.12 & 27.42 & 27.52 & 35.09 & 36.74 & 36.96 & 39.27 & 38.69 & 41.06 & 41.10 \\
ObjectNet & 16.05 & 18.02 & 16.95 & 19.41 & 19.46 & 19.73 & 21.45 & 22.67 & 24.71 & 24.07 & 25.46 & 24.43 & 25.31 & 25.50 & 30.56 & 31.70 & 32.99 & 34.94 & 33.29 & 36.92 & 36.24 \\
Imagenet-A & 7.48 & 9.79 & 8.13 & 10.11 & 9.51 & 10.73 & 11.29 & 12.09 & 14.68 & 13.76 & 15.24 & 14.75 & 14.99 & 15.15 & 18.28 & 21.57 & 19.75 & 24.89 & 21.63 & 24.55 & 25.85 \\
CIFAR-10 & 66.53 & 74.42 & 78.38 & 73.50 & 77.74 & 77.81 & 83.06 & 74.02 & 77.39 & 74.59 & 71.21 & 75.95 & 72.85 & 77.59 & 89.08 & 89.43 & 90.69 & 90.36 & 91.28 & 91.66 & 93.52 \\
CIFAR-100 & 35.38 & 41.87 & 42.19 & 42.27 & 44.60 & 46.03 & 51.75 & 40.86 & 43.47 & 46.21 & 45.58 & 47.84 & 47.51 & 48.72 & 58.88 & 62.07 & 62.53 & 63.87 & 65.69 & 65.21 & 70.17 \\
MNIST & 34.20 & 19.66 & 37.03 & 22.45 & 37.93 & 21.01 & 20.61 & 25.43 & 33.94 & 30.39 & 28.76 & 42.60 & 13.44 & 51.63 & 46.03 & 18.91 & 28.09 & 25.08 & 14.90 & 27.45 & 20.06 \\
Flowers & 15.92 & 19.45 & 19.53 & 23.73 & 20.36 & 19.97 & 22.44 & 19.61 & 19.79 & 20.77 & 25.48 & 20.49 & 24.22 & 21.74 & 22.52 & 29.87 & 25.97 & 32.01 & 29.05 & 32.56 & 29.16 \\
Cars & 2.74 & 4.04 & 3.92 & 5.01 & 4.35 & 5.81 & 4.89 & 4.70 & 5.20 & 4.99 & 5.45 & 4.58 & 4.73 & 4.60 & 16.75 & 19.13 & 18.83 & 22.56 & 21.89 & 23.53 & 23.41 \\
SVHN & 9.22 & 19.81 & 24.48 & 17.03 & 10.98 & 19.46 & 22.12 & 15.41 & 11.66 & 26.28 & 20.69 & 23.97 & 23.42 & 19.50 & 15.21 & 24.83 & 16.59 & 16.61 & 25.96 & 30.71 & 22.76 \\
FER2013 & 24.28 & 26.68 & 30.80 & 28.75 & 34.30 & 30.61 & 40.62 & 18.84 & 32.56 & 31.72 & 41.36 & 34.79 & 34.10 & 28.25 & 24.30 & 34.15 & 25.56 & 38.53 & 35.69 & 30.55 & 41.08 \\
RSST2 & 49.97 & 50.08 & 47.94 & 49.91 & 50.08 & 50.08 & 50.08 & 50.08 & 50.08 & 49.91 & 49.91 & 48.59 & 49.91 & 49.91 & 50.13 & 50.08 & 50.08 & 50.08 & 50.08 & 50.08 & 50.13 \\
Pets & 30.72 & 33.09 & 29.90 & 34.01 & 35.43 & 30.99 & 39.41 & 33.66 & 33.96 & 35.16 & 33.36 & 38.87 & 37.99 & 38.16 & 54.29 & 53.26 & 60.18 & 57.07 & 62.01 & 56.55 & 62.28 \\
Caltech-101 & 66.24 & 70.96 & 71.64 & 70.94 & 71.57 & 73.74 & 73.03 & 70.60 & 71.65 & 71.83 & 71.39 & 72.88 & 72.69 & 72.74 & 77.47 & 78.39 & 77.47 & 78.23 & 78.85 & 79.75 & 79.85 \\
VOC2007 & 43.32 & 40.00 & 46.01 & 46.11 & 52.11 & 51.76 & 53.41 & 40.89 & 44.82 & 48.54 & 47.99 & 46.27 & 50.81 & 51.30 & 58.49 & 64.31 & 71.79 & 55.98 & 67.13 & 65.30 & 70.01 \\
SUN397 & 43.02 & 41.19 & 47.44 & 41.16 & 49.05 & 39.96 & 52.10 & 41.89 & 43.72 & 49.38 & 44.42 & 48.70 & 50.85 & 50.42 & 53.18 & 53.73 & 54.85 & 57.29 & 56.21 & 55.28 & 58.02 \\
Aircraft & 1.74 & 1.23 & 1.38 & 1.83 & 1.68 & 1.77 & 2.01 & 2.01 & 1.11 & 1.38 & 2.01 & 1.32 & 1.38 & 1.56 & 1.95 & 3.45 & 2.07 & 2.70 & 1.83 & 3.81 & 2.55 \\
Country211 & 49.97 & 50.08 & 47.94 & 49.92 & 50.08 & 50.02 & 50.10 & 50.08 & 50.12 & 49.92 & 49.92 & 48.60 & 49.92 & 47.01 & 49.07 & 4.62 & 5.80 & 50.00 & 50.03 & 50.07 & 50.14 \\
DTD & 16.44 & 17.07 & 20.16 & 20.16 & 20.05 & 19.63 & 22.71 & 19.61 & 21.82 & 24.47 & 24.04 & 26.01 & 23.51 & 25.90 & 27.61 & 27.02 & 32.39 & 30.77 & 31.60 & 28.57 & 33.30 \\
GTSRB & 10.26 & 10.15 & 9.25 & 12.72 & 12.04 & 18.31 & 12.91 & 9.95 & 12.19 & 14.51 & 12.91 & 13.26 & 12.68 & 11.80 & 14.66 & 11.68 & 14.06 & 14.50 & 12.76 & 17.14 & 15.64 \\
STL10 & 90.55 & 92.04 & 91.63 & 91.15 & 92.91 & 92.00 & 94.65 & 87.11 & 88.70 & 92.24 & 90.75 & 93.00 & 94.03 & 93.23 & 96.86 & 95.56 & 96.50 & 96.68 & 96.80 & 95.69 & 97.90 \\
Retino & 6.33 & 30.37 & 3.19 & 30.89 & 2.49 & 12.74 & 2.57 & 4.17 & 8.03 & 7.54 & 5.02 & 30.23 & 24.20 & 44.95 & 2.50 & 6.83 & 3.63 & 9.06 & 5.00 & 59.02 & 5.00 \\
EuroSAT & 29.75 & 31.30 & 33.15 & 30.90 & 33.45 & 32.65 & 36.00 & 36.20 & 37.20 & 38.15 & 34.75 & 39.40 & 38.35 & 39.45 & 40.70 & 40.05 & 42.80 & 37.95 & 42.30 & 37.40 & 41.60 \\
RESISC45 & 23.08 & 27.57 & 23.84 & 29.08 & 29.84 & 30.27 & 35.98 & 34.27 & 32.48 & 33.44 & 33.63 & 33.17 & 32.51 & 33.38 & 39.94 & 41.81 & 41.25 & 43.90 & 45.24 & 45.44 & 47.71 \\
PCAM & 40.67 & 49.33 & 55.64 & 57.44 & 49.51 & 54.24 & 48.41 & 53.11 & 54.30 & 55.80 & 54.30 & 55.21 & 65.11 & 49.66 & 57.50 & 54.25 & 51.99 & 54.97 & 50.81 & 48.88 & 55.60 \\
CLEVR Counts & 12.94 & 14.31 & 14.75 & 17.70 & 18.41 & 18.53 & 15.21 & 19.03 & 19.93 & 20.03 & 21.17 & 12.34 & 24.27 & 23.03 & 19.55 & 26.79 & 16.64 & 26.13 & 25.28 & 22.66 & 14.77 \\
CLEVR Dist & 15.87 & 24.43 & 21.47 & 22.61 & 24.14 & 24.81 & 11.43 & 23.10 & 24.05 & 15.75 & 15.93 & 15.83 & 20.15 & 15.80 & 14.93 & 22.67 & 25.38 & 9.42 & 20.77 & 9.98 & 25.79 \\
DSPRITE Orient & 1.84 & 2.97 & 2.52 & 3.32 & 2.42 & 2.76 & 3.20 & 2.51 & 2.83 & 2.47 & 1.96 & 2.89 & 1.90 & 3.03 & 1.13 & 2.48 & 2.40 & 2.59 & 2.46 & 2.65 & 2.74 \\
DSPRITE Position & 3.21 & 3.26 & 3.20 & 3.20 & 3.21 & 3.25 & 3.03 & 3.03 & 3.04 & 3.07 & 3.01 & 3.49 & 3.10 & 3.17 & 3.20 & 3.14 & 3.17 & 3.14 & 3.06 & 3.08 & 3.07 \\
SmallNORB Elv & 11.06 & 9.32 & 10.36 & 10.83 & 13.99 & 10.12 & 11.44 & 11.25 & 12.41 & 10.48 & 11.84 & 11.10 & 10.27 & 10.76 & 11.14 & 13.78 & 11.2 & 10.74 & 11.05 & 10.77 & 10.09 \\
SmallNORB AZim & 4.15 & 6.13 & 5.17 & 5.24 & 5.46 & 5.1 & 4.97 & 5.15 & 5.19 & 6.19 & 6.10& 5.13 & 6.29 & 5.21 & 5.44 & 5.88 & 5.43 & 5.39 & 4.83 & 6.35 & 5.08
\\
DMLAB & 18.06 & 17.04 & 18.01 & 11.96 & 14.40 & 16.67 & 14.49 & 14.25 & 15.75 & 16.64 & 14.61 & 19.07 & 17.44 & 17.73 & 18.21 & 19.43 & 17.20 & 18.29 & 16.16 & 14.15 & 17.69 \\
KITTI Dist & 12.24 & 33.47 & 18.00 & 43.18 & 29.54 & 37.69 & 25.04 & 24.47 & 28.55 & 32.49 & 35.16 & 36.43 & 17.16 & 13.08 & 36.99 & 24.75 & 36.99 & 33.61 & 23.07 & 31.08 & 45.43 \\ \midrule
Avg-35 & 25.61 & 28.84 & 28.61 & 30.08 & 29.89 & 30.39 & 30.78 & 28.72 & 30.55 & 31.35 & 31.17 & 32.53 & 31.80 & 32.68 & 35.35 & 34.97 & 35.22 & 36.98 & 36.89 & 38.90 & 38.89 \\
\bottomrule
\end{tabular}
}
\label{tb:vtab+}
\end{table*}

\begin{table*}[h]
\centering
\renewcommand{\arraystretch}{1.3}
\begin{minipage}{0.54\textwidth}
\renewcommand{\arraystretch}{1.44}
\centering
\caption{\textbf{Zero-shot Top-1 accuracy (\%) on ImageNet-1K for model chains with different expansion ratios.}}
\resizebox{\textwidth}{!}{%
\begin{tabular}{l|ccccccc}
\toprule
 \multirow{2}{*}{\textbf{Model Chain}} & \multicolumn{7}{c}{\textbf{Models}} \\
 & ViT-T/16 & ViT-C/16 & ViT-S/16 & ViT-M/16 & ViT-B/16 & ViT-XB/16 & ViT-L/16 \\ 
\midrule
\textit{baseline} & 26.16 & 28.54 & 30.16 & 31.08 & 31.80 & 32.44 & 33.77 \\ \midrule
2$\times$ Expansion & 26.16 & - & - & 31.22 & - & - & 33.82 \\
4$\times$ Expansion & 26.16 & - & 30.24 & - & 31.83 & - & 34.27 \\
8$\times$ Expansion & 26.16 & 29.05 & 30.32 & 31.46 & 32.14 & 32.76 & 34.33 \\
\bottomrule
\end{tabular}
}
\label{tb:model_chains}
\end{minipage}%
\hfill
\begin{minipage}{0.44\textwidth}
\renewcommand{\arraystretch}{1.0}
\centering
\caption{\textbf{Comparison of Top-1 accuracy (\%) on ImageNet-1K}: baseline vs. CoM-PT under varying training epochs.}
\resizebox{\textwidth}{!}{%
\begin{tabular}{c|c|cccc}
\toprule
\multirow{2}{*}{\textbf{\# Baseline Epochs}} & \multirow{2}{*}{\textbf{Method}} & \multicolumn{4}{c}{\textbf{Models}} \\
 &  & ViT-T/16 & ViT-S/16 & ViT-B/16 & ViT-L/16 \\
\midrule
\multirow{2}{*}{32} & \textit{baseline} & 24.40 & 27.53 & 28.77 & 29.84 \\
                    & CoM-PT   & 24.40 & 28.23 & 29.00 & 29.99 \\
\midrule
\multirow{2}{*}{64} & \textit{baseline} & 25.41 & 28.83 & 30.57 & 32.02 \\
                    & CoM-PT   & 25.41 & 29.49 & 30.75 & 32.03 \\
\midrule
\multirow{2}{*}{128} & \textit{baseline} & 26.16 & 29.98 & 31.80 & 33.77 \\
                     & CoM-PT   & 26.16 & 30.24 & 31.83 & 34.27 \\
\bottomrule
\end{tabular}
}
\label{tab:epochs_comparison}
\end{minipage}
\end{table*}

\subsubsection{The Efficacy of More Sophisticated Designs}
We then compare our best simple approaches with previous sophisticated designs. 
Specifically, we implement Net2Net \cite{chen2015net2net} and NetExpand \cite{ding2023network}, two representative methods for vision model expansion.

Table \ref{exp:ablation_other} presents the comparative results. Our simple implementation achieves performance comparable to these sophisticated designs, delivering 3.76\%$|$5.81\% gains to the baseline on ImageNet-1K$|$COCO, respectively. In comparison, the sophisticated method, NetExpand, only provides a marginal 0.12\% improvement to our approaches on ImageNet-1K.

These results demonstrate that our simple approach is already good enough, which strikes a promising balance between simplicity and performance.

\subsection{Inverse Feature Distillation}
\label{sec:ifd}
In this part, we aim to suggest why we typically ensure $\mathcal{L}_{IFD} < \mathcal{L}_{task}$ by setting an appropriate $\alpha$. Specifically, we train a series of models using CoM-PT with varying coefficients of feature distillation, and observe their performance trend on the validation set of CC3M and ImageNet-1K. 
As shown in Figure \ref{exp:abl_ifd}, feature distillation can enhance the performance of the ViT-S/16 on both datasets when the ratio $r$ between distillation loss and task loss is small, whereas a larger $r$ tends to impair its performance. In the optimal setting, $\mathcal{L}_{IFD}$ is maintained at approximately 10\% of $\mathcal{L}_{task}$, acting strictly as an \textbf{auxiliary loss}. This differs significantly from traditional FD, where the optimal feature distillation loss magnitude is typically larger than the task loss.

\section{More Detailed Results}
\label{sec:e}

\noindent\textbf{Model Performance on VTAB+.}
In Table \ref{tb:vtab+}, we detail the performance of models on each dataset of VTAB+ in the main experiments. From the results, we can see that: i) CoM-PT achieves performance comparable to baseline across most datasets; ii) the average performance across all 35 datasets confirms that CoM-PT maintains lossless acceleration relative to baseline pre-training. Some datasets exhibit inherent instability (e.g., Retino) due to the large domain gaps between the pre-training data and these specialized datasets. However, this instability affects both methods similarly, and the averaged results across the full benchmark provide a robust evaluation that mitigates domain-specific variations.

\vspace{+0.25em}

\noindent\textbf{Model Chains with Different Expansion Ratios.}
To validate our claim in Section \ref{sec:exp_model_chain_design} of the main paper that all models achieve lossless performance, Table \ref{tb:model_chains} details ImageNet-1K results for model chains with varying expansion ratios. The results clearly demonstrate that every model in the chains maintains performance-lossless acceleration compared to the baseline.

\vspace{+0.25em}

\noindent\textbf{Model Performance under Various Training Epochs.} In Table \ref{tab:epochs_comparison}, we illustrate the performance of models pre-trained by baseline and our CoM-PT under varying training epochs. We can observe that our CoM-PT outperforms the baseline on each individual model. This confirms that the acceleration ratios reported in Figure \ref{fig:different_epochs} of the main paper are achieved in a performance-lossless manner.

\end{document}